\documentclass{article}
\usepackage{arxiv}

\usepackage{caption}
\usepackage{natbib}
\usepackage{doi}
\usepackage{xcolor}
\usepackage{color}
\definecolor{darkblue}{rgb}{0, 0, 0.5}
\hypersetup{colorlinks=true, citecolor=darkblue, linkcolor=darkblue, urlcolor=darkblue}
\usepackage{url}
\usepackage{tabularx}
\usepackage{wrapfig}
\usepackage{enumitem}
\usepackage{graphicx}
\usepackage{multirow}
\usepackage{float}
\usepackage{booktabs}
\usepackage{xcolor}
\usepackage{color, soul}
\usepackage{tabularx}
\usepackage{xspace}
\usepackage{makecell}
\usepackage{caption}
\usepackage{subcaption}
\usepackage{array}
\usepackage{float}
\usepackage{listings}
\usepackage{bbm}
\usepackage[ruled,vlined]{algorithm2e}

\usepackage{algorithmic}
\usepackage{tikz}

\usepackage{listings}
\usepackage{color}
\usepackage[T1]{fontenc}
\usepackage{mdframed} 

\SetCommentSty{mycommfont}

\SetKwInput{KwInput}{Input}                
\SetKwInput{KwOutput}{Output}              

\renewcommand{\headeright}{}
\renewcommand{\undertitle}{}
\renewcommand{\shorttitle}{\textit{}}

\hypersetup{
pdftitle={},
pdfsubject={},
pdfauthor={},
pdfkeywords={},
}

\title{Can Foundation Models Help Us Achieve Perfect Secrecy?}

\author{Simran Arora \\
  Stanford University \\
  Stanford, CA \\
  \texttt{simran@cs.stanford.edu}
  \\\And
  Christopher Ré \\
  Stanford University \\
  Stanford, CA \\
  \texttt{chrismre@cs.stanford.edu}\\
}

\newcommand{\systemname}{\textsc{ICL}\xspace}

\usepackage{listings}
\usepackage{color}

\definecolor{dkgreen}{rgb}{0,0.6,0}
\definecolor{gray}{rgb}{0.5,0.5,0.5}
\definecolor{mauve}{rgb}{0.58,0,0.82}

\begin{document}

\maketitle
\begin{abstract}
    A key promise of machine learning is the ability to assist users with personal tasks. Because the personal context required to make accurate predictions is often sensitive, we require systems that protect privacy. A gold standard privacy-preserving system will satisfy perfect secrecy, meaning that interactions with the system provably reveal no private information. However, privacy and quality appear to be in tension in existing systems for personal tasks. Neural models typically require copious amounts of training to perform well, while individual users typically hold a limited scale of data, so federated learning (FL) systems propose to learn from the aggregate data of multiple users. FL does not provide perfect secrecy, but rather practitioners apply statistical notions of privacy --- i.e., the probability of learning private information about a user should be \textit{reasonably low}. The strength of the privacy guarantee is governed by privacy parameters. Numerous privacy attacks have been demonstrated on FL systems and it can be challenging to reason about the appropriate privacy parameters for a privacy-sensitive use case. Therefore our work proposes a simple baseline for FL, which both provides the stronger perfect secrecy guarantee and does not require setting any privacy parameters. We initiate the study of when and where an emerging tool in ML --- the in-context learning abilities of recent pretrained models --- can be an effective baseline alongside FL. We find in-context learning is competitive with strong FL baselines on 6 of 7 popular benchmarks from the privacy literature and a real-world case study, which is disjoint from the pretraining data. We release our code here: \url{https://github.com/simran-arora/focus}
\end{abstract}

\section{Introduction}
A key promise of machine learning is the ability to assist users with personal tasks. Given the private nature of the personal data, these systems should satisfy \textbf{three desiderata: (1) no leakage of private information, (2) quality, and (3) feasibility}. Guided by these desiderata, we propose in-context learning, an emergent capability of recent pretrained models, as a simple approach for privacy sensitive workloads \cite{bommasani2021fm, wei2022emergent}.

The ideal privacy system will offer perfect secrecy: as users interact with the system, the probability that adversaries learn private information should not increase \cite{shannon1949secrecy}. A trivial way to achieve this guarantee is by purely local training or fine-tuning a public model on a user's private dataset. 
However, recent neural models require copious amounts of training data \cite{dodge2020smallfinetune} and 
users often hold a limited scale of labeled data. On average across our evaluation tasks, an individual user has 149 (std. dev. 191) labeled training examples. To address the challenge that individual users lack sufficient data, federated learning (FL) over data spanning multiple privacy scopes (i.e. users) has emerged as a popular approach \cite{shokri2015privacy, mcmahan2016fl}. Indeed \citet{wu2022motley} finds that the local training baseline performs up to 60\% worse than FL on average across users.

\begin{figure*}
    \centering
    \includegraphics[width=0.84\linewidth]{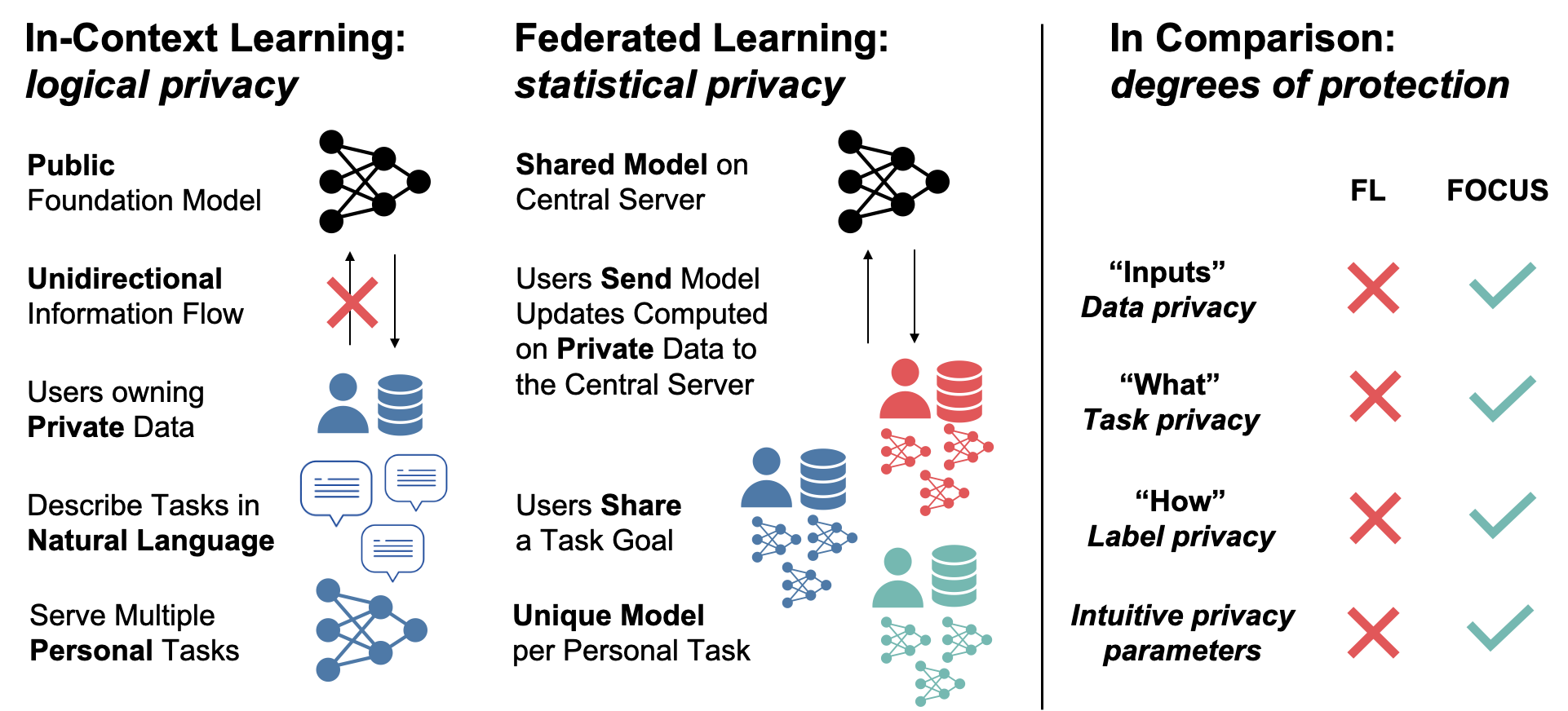}
    \caption[width=\linewidth]{Towards the goal of providing users with perfect secrecy guarantes, we propose in-context learning as a privacy baseline. The proposal entails shipping FMs to users and using in-context learning to adapt to personal tasks.}
    \label{fig:main}
\end{figure*}

Since requiring all users to send data to a central location sacrifices data privacy, FL trains a task model by sending \textit{the model} between users and a central server. In each training iteration, (1) the model is sent to each user, (2) users compute gradients on their local data and send the results back to the central server, where (3) they are aggregated to update the central model. FL has gained widespread popularity as an approach to privacy. Evaluating FL along the desiderata: 
\begin{itemize}
    \item \textbf{Privacy} 
    FL does not require users to upload raw private data, however private information can be recovered from the exposed model \cite{shokri2017meminference, melis2019flleakage, nasr2019meminference, bagdasaryan2020backdoorfl}. To improve the privacy guarantee over the training inputs, a popular strategy is to insert statistical noise to the training procedure \cite{dwork2006dp}. Inserting more noise improves the privacy guarantee, but can degrade quality. 
    The world's largest organizations struggle to reason about how to set the appropriate privacy parameters \cite{greenberg2017wired, mcmahan2018dpfl}. 
    \item \textbf{Quality} 
    Standard FL algorithms improve model performance for the \textit{average} user. However private data often widely differs by individual and performance fluctuates across participants \cite{zhao2018noniid, wu2022motley}. Personalized FL is a popular research area aimed at addressing this challenge \cite[inter alia.]{arivzhagan2019fedper, li2021ditto}, but can be brittle in non-IID settings. \citet{wu2022motley} demonstrates that even under personalized FL algorithms, up to 40\% of clients experience \textit{worse} performance than if a non-personalized FL algorithm were used on certain tasks. Finally, adversarial participants can corrupt the FL procedure \cite{yin2018byzantine}.
    \item \textbf{Feasibility} Executing FL requires rounds of communication between multiple users \cite{knecn2016fl}.
    The protocol must be repeated for \textit{every personal task} a user wishes to perform. FL also requires aligning the incentives of users who share task goals and may exclude users who cannot contribute training data \cite{zhan2021incentives}.
\end{itemize}

Given the challenges and opportunities of FL, we revisit the question of when users should apply FL to their privacy sensitive workloads. Specifically, we observe natural compatibilities between the capabilities of recent pretrained models, referred to in the literature as Foundation Models (FMs) \cite{bommasani2021fm}, and the above challenges. Traditionally, the models are adapted to downstream tasks through a task specific fine-tuning step in which all model weights are updated. Recent FMs often provide impressive quality on new downstream tasks out-of-the-box or with minimal effort, simply given a textual description of the task and zero-to-few task demonstrations, which is known as \textit{in-context learning} (ICL) \cite{brown2020language, radford2021clip}. ICL is applied at inference time; no additional training is required. Interestingly, ICL has proven effective in tasks that are widely different from the objective for which the FM was originally trained \cite{chowdhery2022palm, liu2022commonsense}. 

In this work, we observe that ICL can provide perfect secrecy, in contrast to FL, and therefore investigate how the performance and systems feasibility compare to FL. While prior work compares \systemname{} to other non-private methods of adapting FMs to downstream tasks, such as lightweight or full fine-tuning \cite{brown2020language, liu2021promptsurvey}, we initiate the study of \systemname{} as it compares to federated learning. 
Our proposed architecture entails sending off-the-shelf public FMs to private user silos and locally applying in-context learning to perform personal privacy-sensitive tasks. 
We initiate the study of when and where this simple baseline may be effective by studying $9$ popular FMs and $7$ popular evaluation settings in the privacy literature, which span  language and vision.
\begin{itemize}
    \item \textbf{Privacy: What privacy guarantee does \systemname{} provide?} \systemname{} provides perfect secrecy given the unidirectional information flow: FMs are sent to the user, no information leaves the user silo. Formally, \systemname{} uses an access control based privacy framework \cite{bell2976blm}.
    Beyond data privacy, we observe \systemname{} offers \textit{Task-Privacy}. In FL, users share the same task goal and label schema in order to train a central model, but \systemname{} hides \textit{what} the user is doing and \textit{how} they are doing it (i.e. the label schema for the task). 
    \item \textbf{Quality: Is \systemname{} competitive with FL?} Privacy tasks have not previously been benchmarked with ICL techniques. We initiate this benchmarking and find the \systemname{} baseline is competitive on 6 of 7 tasks from the privacy literature. However, FL performs much better on fine-grained tasks and when the inductive bias of the FM is misaligned with the task. We also evaluate on a real-world information extraction task on data that is disjoint from the FM pretraining corpus.
    \item \textbf{Feasibility: What are the systems costs of \systemname{}?} We characterize the landscape of costs in terms of (1) the training and inference costs: \systemname{} eliminates the costs of training and communicating the model for multiple iterations, but incurs large inference costs; and (2) the number of users who need to share the same task goal in order for FL to succeed: under \systemname{}, users do not need to share the same task objectives. Finally, the FL literature traditionally reports the cost of supporting a \textit{single} task. Yet, users may want to perform \textit{multiple} private tasks over their data (e.g., classification and QA). We propose practitioners should incorporate the \textit{number of tasks} we can support with a single model when evaluating our privacy baselines --- one FM can support multiple tasks.
\end{itemize}
In summary, we initiate the study of in-context learning with recent FMs as a baseline for FL. We conduct an extensive empirical analysis of the proposed baseline and hope this encourages further work on how we can provide users with perfect secrecy guarantees for their sensitive tasks. 

\section{Background and Related Work}
We focus on performing personal tasks over privacy-sensitive data. Complementary work studies how to release FMs that have been trained over sensitive data \cite{tian2022fedbert, li2022dpllm}, which is beyond our scope. We further focus on the setting where users have a limited amount of personal data as in \cite[inter alia.]{caldas2019leaf, dodge2020smallfinetune, wu2022motley}.

\textbf{Perfect secrecy} 
A system preserves perfect secrecy if the probability of an adversary obtaining knowledge of a client's private data does not increase as the client interacts with the system \cite{shannon1949secrecy}. As the system performs multiple tasks over the same underlying private data, the probability an adversary obtains knowledge of the private data should not increase. Mathematically, these are \textit{logical notions} of privacy \cite{miklau2004secrecy}.

\textbf{Machine learning for personal tasks} The above properties are achievable through training or fine-tuning on only the user's private data, for which existing work assumes users own a sufficient amount of private labeled training data \cite{xu2018deeptype}. However, fine-tuning is found to be unstable in the low-data regime \cite{dodge2020smallfinetune, mosbach2021smallfinetune, wu2022motley}. Therefore federated learning (FL) proposes to train a model over the private data owned by multiple parties \cite{mcmahan2016fl}. FL initializes a global model $\theta$ and solves:
$$\min_{\theta} G(L(\theta;D_1), ..., L(\theta; D_i), ..., L(\theta; D_N))$$
where $L(\theta;D_i)$ are the local objectives for each client, and $G(\cdot)$ is a function of the local objectives. In training, each client receives a copy of model $\theta_t$ at the current timestep $t$ from a central server, performs gradient descent on the local data $D_i$, and sends the gradients to the central server, where the local gradients  are combined using $G(\cdot)$ to produce $\theta_{t+1}$. FedAvg \cite{mcmahan2016fl}, a weighted sum of the local objectives, is the vanilla choice for $G(\cdot)$. \citet{li2019flsurvey} and Appendix \ref{sec:appendix_baselines} describe other objectives. 

To achieve personalization in FL, popular approaches include using a mixture of local and global model updates, training a global model with model-agnostic meta-learning so it may be quickly adapted to local data, and initializing the global model with a pretrained initial model \cite{fallah2020fedmaml, zhao2018noniid, hikmil2021fedfinetune, wu2022motley}. These methods improve upon training from scratch with FedAvg, though do not provide perfect privacy. Empirically, the improvements of these methods over FedAvg also appear to degrade in increasingly non-IID settings \cite{chen2021fedmatch, arivzhagan2019fedper}. \citet{wu2022motley} reports that up to 40\% of clients can experience \textit{worse} performance under personal FL compared to non-personal FedAvg on certain tasks.

The key constraint under FL is that the raw data $D_i$ never leaves the private silo for client $i$, however, functions of $D_i$ may be exposed. Unfortunately, attacks on the gradients or inference attacks on the final model output can reveal private information \cite{shokri2017meminference, melis2019flleakage}. Techniques to protect the privacy of the communicated parameters include differential privacy (DP) \cite{dwork2006dp}, which degrades model performance \cite{mcmahan2018dpfl}, or secure multiparty computation \cite{bonawitz2017smpc}, which does not protect against the membership inference attacks and significantly increases the computational cost \cite{fang2021smpc}.

\textbf{Foundation models} Foundation models (FMs) are trained via self-supervised objectives on large corpora of publicly available data (e.g. web pages or image-caption pairs). Traditionally, the models are adapted to downstream tasks through a task specific fine-tuning step in which all model weights are updated. Impressively, recent FMs can solve new tasks by conditioning their predictions on natural language specifications of the task, with no additional training. This is referred to as \textit{in-context learning} (\systemname{}).

We are certainly not the first to use public models to achieve improved privacy guarantees \cite{kerrigan2020dplm, yu2022dpfl, li2022dpllm}. The prior works use the FM as an initialization and adapt the model to downstream personal tasks via federated fine-tuning. We initiate the study of \systemname{} as a baseline for privacy-preserving ML and evaluate whether \textit{perfect secrecy} is feasible using \systemname{}.

\section{In-Context Learning Baseline}
\label{sec:method}
In this section, we introduce the in-context learning baseline.

\subsection{Baseline}
Consider client $i$ who wants to perform several personal tasks $t \in T_i$ over their private data.
For $t$, the client has a private dataset $D_{it} =  \{x_j, y_j\}_{j=1}^{j=n_i}$, where $n_i \in [0, k]$ for some small $k$. The \systemname{} architecture centers on a \textit{unidirectional dataflow}, as depicted in Figure \ref{fig:main}. The user downloads the FM to the private silo and uses in-context learning to perform the private task.

\paragraph{In-context learning} Language FMs are pretrained to learn a probability distribution $p_\mathrm{FM}(x)$ over large coprora of text $x$, drawn from the distribution $p$, such that $p_\mathrm{FM} \approx p$. Popular LMs are factorized as $p_\mathrm{FM}(x) = \prod_{i=1}^{|x|} p_\mathrm{LM}(x_i|x_{< i})$. The FM is tasked with predicting the next token $x_i$ given the context $x_{< i}$. Recent FMs have been trained on large amounts of diverse data, and demonstrate the ability to transfer to new tasks simply given natural language descriptions of the task \cite{brown2020language, wei2022emergent}. This inference-time adaptation is termed \textit{in-context learning}.

Current text-image FMs are pretrained over pairs of images and paired text data in the same latent representation space. Given pairs $D_{paired} = \{(v_i, x_i)\}_{i=1}^{n_{paired}}$, for image $v_i$ and text $x_i$ drawn from $p_{\mathrm{FM}}$, the FM learns embedding functions $f_{image}$ and $f_{text}$ such that the distances between the embeddings $\mathrm{dist}(f_{image}(v_i), f_{text}(x_j))$ reflect the semantic similarity between $v_i$ and $x_j$ for all $i, j$. During inference, the user can design contexts $x_1, ..., x_k$ (for instance representing different classes in a classification task) and given a new image, determine the most similar text (class).

\paragraph{Designing the context} The context is defined by a template, which contains placeholders for the task description and demonstrations of the inputs and outputs for the task. Context design is a brittle process --- small changes to the template and choice of demonstrations can result in large performance variations \cite{zhao2021calibrate} --- so significant recent work studies how to write effective contexts \cite[inter alia.]{mishra2021reframing, wu2022chains, wei2022cot}. In this work we use standard task-agnostic templates --- we randomly select demonstrations from the user's local dataset $D_{it}$ for task $t$ --- to understand the performance of ICL without over-engineering the contexts.

\subsection{Properties}
The ideal baseline for solving personal tasks over privacy-sensitive data should provide strong privacy guarantees and high quality along ML metrics. Here we define privacy model of our proposal and analyze when we expect the \systemname{} baseline to be effective.

\paragraph{Privacy guarantee}  \systemname{} provides access control privacy with perfect secrecy ($\epsilon = 0$) since no private data leaves the user device \cite{bell2976blm, hu2006nist}. In our setting, the FM is owned by public entities and users own zero-to-few private task demonstrations and task descriptions, which remain on device. No private data or function of private data is exposed to untrustworthy subjects. FL violates this rule by exposing functions of the private data to untrustworthy entities. Untrustworthy entities cannot send data to the user under our framework. FL violates this by allowing the central server and other users to tamper with the data incorporated in the user's local model. 

Additionally \textit{task privacy}, which encompases the task (e.g. sentiment classification or QA) the user is performing and the label schema (e.g. binary or 5-class sentiment classification) immediately emerge from \systemname{}, while neither is protected under FL.

\paragraph{Quality} \systemname{} is expected to be compelling in settings with the following data and task properties: 

\begin{enumerate}
    \item \textbf{Non-IID user data} FL is challenging over non-IID data and  requires users to share the same task goals, sacrificing task privacy. With \systemname{}, the client task and data distributions are independently handled.
    \item \textbf{Public and user distributions} \systemname{} is likely effective when the sensitive tasks of interest are well-represented in the public pretraining data. The degree to which sensitive tasks are reflected in pretraining is an open-question. 
    \item \textbf{Data scale} A client's training dataset size, $n_i$, is small enough that training or fine-tuning purely locally results in low-quality models \cite{dodge2020smallfinetune, mosbach2021smallfinetune, wu2022motley}. \systemname{} uses $0-k$ task demonstrations to guide the FM to perform new tasks.
\end{enumerate}

\section{Experiments and Analysis}

To study the benefits and limitations of FMs for privacy, we ask the following questions:
    (1) Is \systemname{} competitive in quality to leading privacy frameworks? How does this vary by FM size and bias, and task properties?
    (2)  To what degree does \systemname{} enable personalization to the user's data distribution? 
    (3) Is \systemname{} effective on data that is disjoint from the pretraining distribution? 

\subsection{Experimental Setup}
\label{sec:experiments}
\textbf{Benchmarks} We use a representative set of standard benchmarks in the privacy literature, which are useful proxies for personal tasks such as intent and content classification, message completion, and QA \cite{caldas2019leaf, chaoyanghe2020fedml, fedscale-arxiv, fednlp2021}. Each benchmark contains a small number of examples per user with mean 149, standard deviation 191 examples across users and tasks. 

Our tasks include: \textbf{Text classification}  Sentiment140 2-way \cite{go2009sent140} and 20News 20-way classification benchmarks \cite{lang1995news20}, \textbf{Image classification} CelebA binary \cite{liu2015celeba}, CIFAR10 10-way \cite{krizhevsky2009cifar10}, and Federated EMNIST 62-way classification benchmarks \cite{cohen2017emnist},  \textbf{Language modeling}  Reddit benchmark \cite{caldas2019leaf},  \textbf{Reading Comprehension}  MRQA benchmark \cite{fisch2019mrqa}.

\paragraph{Foundation models}
\label{sec:baselines}
We use a representative set of FMs including T0 (3B and 11B parameters) \cite{sanh2022t0},  and GPT-3 (125M, 1.3B, 2.7B, 6.7B and 175B p.) \cite{brown2020language} for prompting tasks, MPNet-base bi-encoders for textual zero-shot similarity-search (110M p.) \cite{reimers-2019-sentence-bert, song2020mpnet}, and  CLIP for image zero-shot similarity-search (150M p.) \cite{radford2021clip}. 

We characterize the above models by their pretraining strategy. T0 was fine-tuned on pairs of context-input-output tuples to improve its ICL preformance. On the other hand, GPT simply uses next word prediction and no  fine-tuning. 

The above models report the pretraining data and we confirm that the private benchmarks are not included during pretraining. \footnote{However for Reddit, most FMs are trained on generic Common Crawl, which contains Reddit data. We thus
use Grover \cite{zellers2019grover}, which are pretrained only on news articles and find their result directly match or exceed the performance of GPT models of the same size --- 9.6\% for GPT-125M vs. 11.4\% for Grover-124M, and 13.2\% for GPT-1.3B vs. 13.4\% for Grover-1.5B. Due to the limited range of Gover sizes and given the comparable quality of both models, we proceed with GPT variants.} We also release our code and prompts.

\begin{table*}[t!]
        \begin{center}
    \normalsize
    \begin{tabular}{llccc}
    \toprule
      \multirow{2}{*}{\textsc{Benchmark}}  &      \multicolumn{2}{p{5cm}}{\centering \textsc{In-Context Learning}} &      \multicolumn{2}{p{4.5cm}}{\centering \textsc{Federated Learning}}  \\ 
         &  \emph{Method} &   \emph{Accuracy}  &  \emph{Method}  & \emph{Accuracy}   \\
         \midrule

    Reddit & 
    Prompting (175B)
    &  \textbf{13.6} & FedAvg (1M) \cite{caldas2019leaf} & 13.4 \\
    
    MRQA* & 
    Prompting (175B)
    &  \textbf{64.1} & FedAvg (67M) \cite{fednlp2021} & 27.1 \\
    
    Sent140 & 
    Prompting (175B)
    & \textbf{80.4} & FedAvg (1M) \cite{michieli2021sent140baseline} & 69.5 \\
    20News &  
    Prompting (175B)
    &  31.3 & FedAvg (67M) \cite{fednlp2021} & \textbf{51.4} \\
    
    \midrule
    Sent140 & 
    \textit{Prompting} (11B)
    & \textbf{78.2} & FedAvg (1M)  \cite{michieli2021sent140baseline} & 69.5 \\
    20News & 
    \textit{Prompting} (3B)
    &  19.1 & FedAvg (67M) \cite{fednlp2021} & \textbf{51.4} \\
    
    Sent140 & 
    \textit{Similarity-search} (110M)
    & 61.5 & FedAvg (1M)  \cite{michieli2021sent140baseline} & \textbf{69.5} \\
    
    20News & 
    \textit{Similarity-search} (110M)
    &  \textbf{63.4} & FedAvg (67M) \cite{fednlp2021} & 51.4 \\
    \midrule
    CelebA  & 
    \textit{Similarity-search} (150M)
    & \textbf{86.7} & FedAvg (22M) \cite{qu2022vitfl}  & 86.6 \\
    CIFAR10 & 
    \textit{Similarity-search} (150M)
    &
    \textbf{88.2} & FedAvg (22M) \cite{mcmahan2016fl} & \textbf{98.5}/86.9** \\
    F-EMNIST  & 
    \textit{Similarity-search} (150M)
    & 22.3 & FedAvg (45M) \cite{rothchild2020fetchsgd} & \textbf{81.0}  \\
    
    \bottomrule
    \end{tabular}
    \normalsize
    \caption{Test results on standard privacy benchmarks using FMs in \textbf{zero-shot} adaptation. \textit{Italics} indicates the FM was fine-tuned on task or instruction-label pairs after pretraining. *F1 score. **Trained from scratch.} 
    \label{tab:benchmarks}
    \end{center}
\end{table*}

\paragraph{Federated learning baselines}
We report FL numbers where FL is initialized with SoTA pretrained architectures and task adaptation is performed using FedAvg. FL is initialized with ViT(S) (22M p.) and ResNet101 (45M p.) for vision tasks, DistillBERT for 20News and MRQA (67 M p.), and Stacked-LSTM for Sent140 and next-word prediction (1M p.). 
FedAvg is performed over 10k-1Ms of examples, 5-1000s clients and 10-1000s rounds of model communication for each task \cite{chaoyanghe2020fedml, caldas2019leaf, fednlp2021, fedscale-arxiv}.

\paragraph{Model selection} An extended discussion of how the \systemname{} and FL baselines are selected is in Appendix \ref{sec:appendix}, and a concrete memory and computation comparison is in Section \ref{sec: systems}. Overall, there is a large search space for designing the context in in-context learning \cite[inter alia.]{wu2022chains, mishra2021reframing, wei2022cot}, and similarly in FL, there several training  objectives have been proposed.
We compare \systemname{} to FL baselines that apply no differential privacy or protections against adversarial parties, or for certain benchmarks, client heterogeneity, all of which degrade performance \cite{mcmahan2018dpfl, li2021ditto, rothchild2020fetchsgd, sun2021decentralizedfl}. 

Given the many potential variations to each baseline, we compare the standard implementations for each. The purpose of the evaluation is to uncover which portions of the privacy-quality-feasibility tradeoff space the baselines occupy and not to conclude one baseline is strictly preferable.

\subsection{Does \systemname{} give competitive  quality?}
First we evaluate the performance of \systemname{} using zero-shot prompts, i.e. the context contains a task description and no task demonstrations.  Results in Table \ref{tab:benchmarks} show that in the zero-shot setting, FM performance competes with FL performance on 6 of 7 benchmarks. Critically, the above \systemname{} numbers provide perfect secrecy and the FL numbers provide no privacy. We next analyze \textit{when} \systemname{} is competitive based on the FM size, task-difficulty, and applied prompting strategy.

\begin{figure*}
    \centering
    \includegraphics[width=0.8\linewidth]{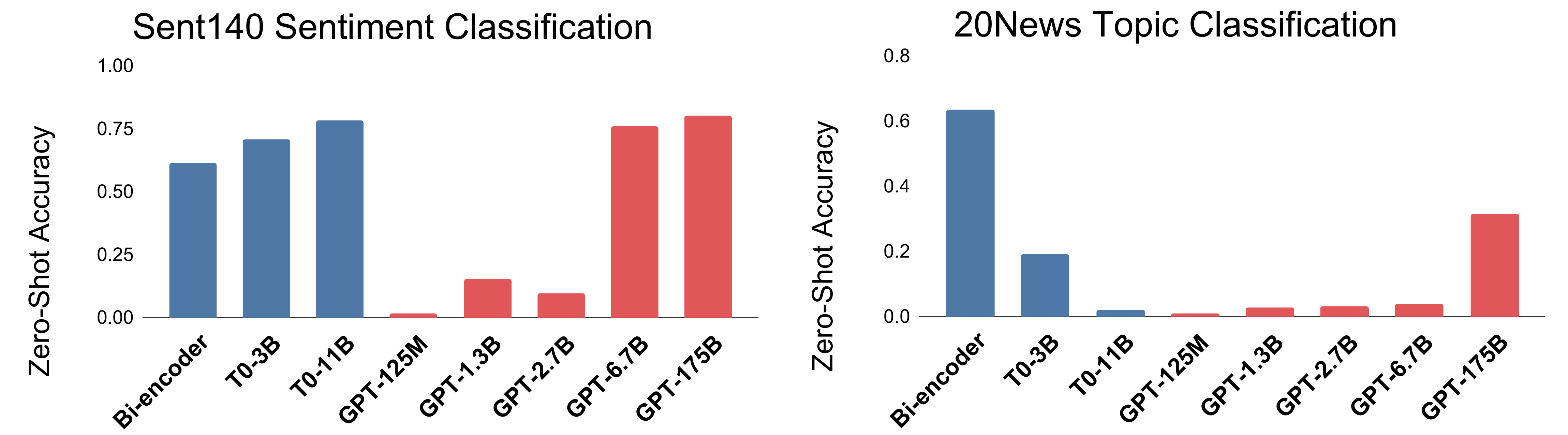}
    \includegraphics[width=0.8\linewidth]{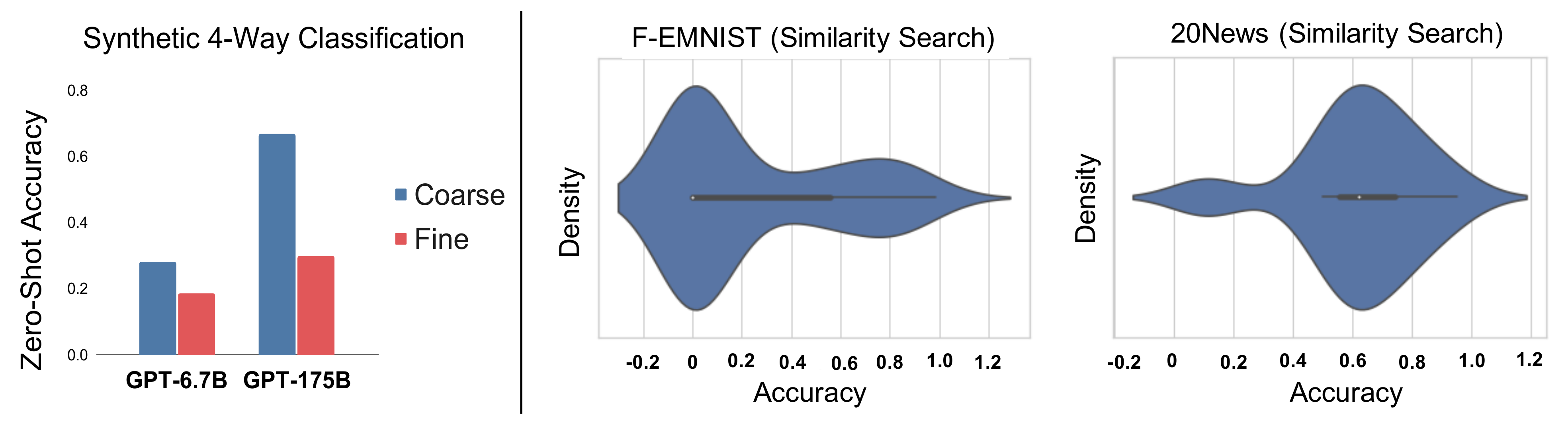}
    \caption[width=\linewidth]{\textbf{Top}: Zero-shot performance by model size and type. Blue indicates ICL fine-tuned FMs and Red indicates non fine-tuned FMs.
    \textbf{Bottom}: FMs struggle on fine-grained classification (left) and can provide highly uneven zero-shot performance across classes (right). Additional violin plots by \textit{user level} accuracy are in Appendix \ref{sec:appendix_results}.}
    \label{fig:sizes}
\end{figure*}

\paragraph{Model size} Larger FMs reliably provide higher quality (Figure \ref{fig:sizes}). Figure \ref{fig:sizes} shows FMs that were fine-tuned for ICL in blue and FMs without fine-tuning in red. Fine-tuning \textit{can} drastically improve performance with orders of magnitude fewer parameters (e.g., on 20News with bi-encoders), however, when there is misalignment between the fine-tuning and the downstream tasks, these FMs fail. In particular, inspecting T0-3B FM mispredictions on 20News, for 24.4\% of examples, the model generated one of the valid classes for a particularly similar topic classification task called AGNews, which, unlike 20News, \textit{was} part of the T0 training data. The issue is far more severe for T0-11B, which entirely fails on the task. This never happens for GPT mispredictions. ICL fine-tuning over more data may help maintain the model generality.

\paragraph{Task-difficulty} FMs entirely fail on the fine-grained classification tasks, F-EMNIST and 20News, which are relatively disjoint from the public pretraining data.
To study this, we create 4-class synthetic classification tasks using subsets of 20News. The \textit{Coarse} synthetic requires classifying an input $\in$ \{politics, automobiles, baseball, or medicine\}, while the \textit{Fine} synthetic requires classifying an input $\in$ \{politics, politics of guns, politics in the Middle East, or religion\}. 
GPT-175B incurs a 55\% drop from Fine to Coarse (Figure \ref{fig:sizes} (left)). 
Further, zero-shot performance is quite uneven across classes on F-EMNIST and 20News (Figure \ref{fig:sizes} (right)) and across users with non-IID private distributions (Figure \ref{fig:violin}, Appendix \ref{sec:appendix_results}), suggesting FMs are not robust zero-shot.

\paragraph{Prompting Strategy} We finally implement and evaluate two advanced \systemname{} methods beyond zero-shot prompting to study the opportunities for improved performance under ICL, which include:
\begin{itemize}
    \item \textbf{Prompt-decomposition} For tasks with large label-spaces, we decompose the task such that the model must decide between a smaller number of label choices in each prompt, to reduce task-difficulty. For instance in the 20-Way News task, we can (1) decompose the label space into 5 groups of 4 labels each at random, (2) prompt the FM 5 times to output the 5 respective labels of each 4-way classification, and (3) prompt the FM to output the answer from the shortlist of 5 labels. 
    \item \textbf{Prompt-calibration} \cite{zhao2021calibrate}. The motivation for this approach is that the initial FM might be biased towards certain output classes and we can use Platt scaling to estimate and adjust for this bias. Given the raw FM output probabilities per class, $p$, we can adjust using the equation: $q = \textrm{softmax}(Wp + b))$.  The method involves taking the FM’s token-probabilities on “content-free inputs” (e.g. passing in the empty string to the FM), and setting the values for $W$ and $b$ such that the outputs for the label-tokens have a uniform probability distribution. Biases can arise because some label tokens are more common than others. 
\end{itemize}

\begin{table*}[t!]
        \begin{center}
    \normalsize
    \begin{tabular}{lcc}
    \toprule
      \multirow{1}{*}{Privacy Benchmark}  &  Method  & Accuracy \\
    \midrule
     \multirow{3}{*}{20News} & Zero-Shot Baseline   & 31.3  \\
     & Prompt-Decomposition  &  48.1 ($+ 53.7\%$) \\
     & Prompt-Calibration \cite{zhao2021calibrate} &  58.0 ($+ 87.1\%$) \\
    \bottomrule
    \end{tabular}
    \normalsize
    \caption{Results using advanced FM prompting methods within the \systemname{} framework on standard privacy benchmarks. These results use the GPT3-175B FM, each on a random set of 100 test examples.}. 
    \label{tab:advanced_LLM}
    \vspace{-7mm}
    \end{center}
\end{table*}

Results from applying the two methods are in Table \ref{tab:advanced_LLM}. 
On 20News, prompt-decomposition results in a 53.7\% improvement over the naive zero-shot FM baseline in Table \ref{tab:benchmarks} (31.3\%). This brings performance to 3.3\% points below the FL baseline. Meanwhile, applying the calibration method (with no additional decomposition) results in an 87.1\% improvement over the standard FM baseline,  which is 6.6\% \textit{above} the FL baseline in Table \ref{tab:benchmarks}.

\subsection{To what degree does \systemname{} allow personalizing to the user's data distribution?} 
Personalization is a well-studied question in the FL literature \cite{arivzhagan2019fedper, fallah2020fedmaml} and we similarly evaluate \systemname{} along this axis. Each user often holds a few training examples per task which are used for personalization. We consider users with $< k$ and  $0$ train examples, for some small $k$.

\begin{figure*}
    \centering
    \includegraphics[width=\linewidth]{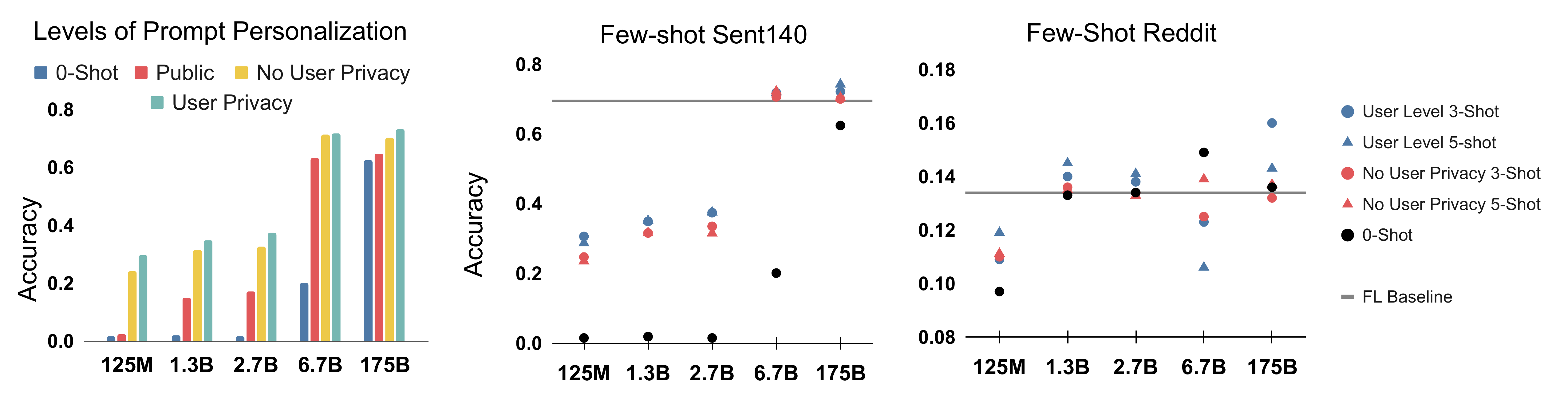}
    \caption[width=\linewidth]{\textbf{Left}: Sent140 results including the ``Public Prompt'' baseline. 
    \textbf{Right}: Reddit and Sent140 results comparing 0-shot vs. k-shot performance by k, and comparing k-shot with examples selected from the user's small labeled training dataset (``User Privacy'') vs. randomly from the aggregate training dataset of all users (``No User Privacy''). Accuracy is averaged across users.}
    \label{fig:gptreddit}
\end{figure*}

\paragraph{Users with few $< k$ private examples.} Now we include task demonstrations in the context for ICL. We compare: randomly chosen training examples from the aggregate pool of all users' training data (``No User Privacy'') and examples chosen from the few training examples the user has available (``User Privacy''). The context includes \textit{no} task description to focus on the effect of task demonstrations. 
We report results for $K \in \{0, 3, 5\}$ on Reddit and Sent140 in Figure \ref{fig:gptreddit}.

We observe: (1) $K>0$ task demonstrations provides improvements over the zero-shot baseline across model sizes, barring GPT-6.7B on Reddit. (2) Few-shot learning enables small FMs to match the performance of FMs with orders of magnitude more parameters --- for example, on Sent140, comparing GPT-125M k-shot performance to GPT-6.7B 0-shot, and GPT-6.7B k-shot to GPT-175B 0-shot. (3) We observe personal user-level demonstrations consistently outperform non-personal contexts.

\paragraph{Users with $k=0$ private examples.} Several personal tasks (e.g., spam classification, message generation) are of interest to many users. We suggest that a \textit{public prompt repository}, or library of public k-shot examples, can be shipped down to users along with the FM. \footnote{Excitingly, relevant resources are already in development, originally proposed for alternate purposes \cite{bach2022promptsource}.} Accordingly, we evaluate ``Public'' third baseline beyond ``No User Privacy'' and ``User Privacy'', where we use hard-coded task demonstrations for every inference example and user. As a proof of concept, Figure \ref{fig:gptreddit} (left) demonstrates that this provides a clear boost over zero-shot FM performance.

Overall, there are clear improvements from the ``Public'' to ``No User Privacy'' to ``User Privacy'' baseline, which is intuitive from a distribution shift perspective. The FMs are sensitive to small peculiarities across users, demonstrating the ability to incorporate personal information during inference. A few in-context examples enable using orders-of-magnitude smaller FM. We include an extended analysis of why user-level examples may help in Appendix \ref{sec:appendix_results}.

\begin{figure}
    \centering
    \includegraphics[width=0.4\linewidth]{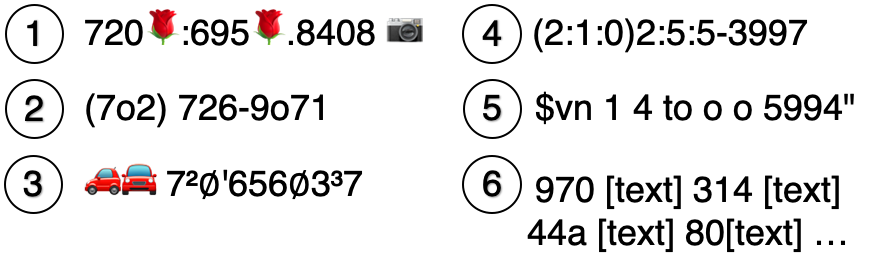}
    \caption[width=\linewidth]{Sample phone numbers embedded in the ads.}
    \label{fig:phones}
\end{figure}

\subsection{Real world case study: Does \systemname{} support out-of-distribution sensitive tasks?}
Thus far, we have evaluated \systemname{} on traditional privacy benchmarks. A pessimistic view of the results is that the canonical privacy benchmarks are not representative of sensitive tasks that users care about. While it is an open question as to the proportion of sensitive tasks that are represented during pretraining, here we demonstrate a case study on real-world sensitive data analysis tasks. We observe that ICL is effective on tasks that are clearly disjoint from the data the FMs saw during pretraining \cite{gao2021pile}.

\paragraph{Out-of-Pretraining-Distribution Data} We evaluate two tasks: 1) phone-number and 2) incall and outcall price information extraction from highly noisy human-trafficking ads. The tasks are critical data-analysis steps for legal entities and researchers seeking to investigate human-trafficking patterns towards its prevention. The advertisements were scraped from the website Backpage.com \cite{delateur2019backpage}, that "hosted more than 80 percent of the online advertising for illegal commercial sex in the United States" \cite{doevbackpage2016} until it was shut down by the Department of Justice in 2018. The data is highly \textit{adversarial} as advertisers use obscure and code language to hide from law-enforcement. We include examples of phone-numbers as they appear in Figure \ref{fig:phones}. 

\paragraph{Results} We evaluate on manually labeled evaluation sets of 100 ads for phone number extraction and 128 ads for  price extraction. As a baseline, the data analysts spent significant effort designing regular expressions and using this, their pipeline yields 61.4\% on phone numbers and 58.6\% of the prices. Using two fixed task demonstrations,  GPT3-175B achieves an 94.1\% accuracy on phone number extraction and 86.6\% accuracy on price extraction. GPT3-6.7B achieves 74.3\% on phone and 66.9\% on price extraction.

\paragraph{Analysis} Though the ad data is disjoint from the pretraining data, ICL is effective and requires low effort (simply two task demonstrations). Further, under FL, the analysts would need to \textit{repeat} the procedure twice to train the phone extraction and the price extraction models. Moreover consider if one analyst \textit{only} is interested in phone extraction, the second is only interested in prices, while a third is interested in a custom extraction. For each of these tasks of interest, under FL, the analysts would need to assemble cohorts of other analysts with the same task goals to obtain the model. Overall, the upfront cost is high with FL in this setting. 

However, the 175B FM far outperforms the 6.7B FM, and acquiring the resources to host 175B models may be challenging. Further, phone numbers and prices are intuitively concepts that are well-represented during pretraining, even though the noisy phone numbers and prices from the ads are not found. We also see that on F-EMNIST and 20-News in Table \ref{tab:benchmarks}, FL significantly outperforms \systemname{}. Overall, an important question to investigate in future work is whether existing ML benchmarks are appropriate for measuring the ability of FMs to transfer to sensitive domains. It is challenging to anticipate when and why FMs underperform and whether sensitive tasks that are of interest to users are well-represented in the  corpora used to pretrain popular FMs.

\section{Systems Feasibility}
\vspace{-2mm}
\label{sec: systems}
We next examine the feasibility of the FL and \systemname{} baselines. Additional discussion can be found in Appendix \ref{sec:appendix}.

Systems costs under FL and \systemname{} are shown in Table \ref{tab:communication}, using an order 100M parameter model for FL and 10B parameter model for \systemname{}, based on the trends in Section \ref{sec:experiments}.  \textbf{Communication} \systemname{} requires locally downloading and FL requires communicating the model repeatedly between users and the central server.  \textbf{Training} requires $(2 \times 3 \times \textrm{model parameters} \times \textrm{steps} \times \textrm{batch size} \times \textrm{input length})$ FLOPs, and communication between clients and the central server, a function of network download/upload speeds and model size. FMs in \systemname{} are frozen. \textbf{Inference} takes $(2 \times \textrm{model parameters} \times \textrm{input length})$ FLOPs,  
assuming key and value vectors for the attention computations are cached.  As we freeze and perform \textit{inference} with FMs, we can use quantization (i.e., 8 or even 4-bit precision) and pruning, to improve efficiency. 

Critically, FL incurs these costs \textit{per personal task} the client cares about. Meanwhile, it is possible for a single FM to transfer to multiple tasks given different ICL contexts. The number of tasks served with the same model is an important factor in evaluating the feasibility of the baselines.

\vspace{-2mm}
\paragraph{Analysis} The limited RAM on modern phones is a key challenge for using large FMs on device today. Future-looking, we are optimistic, noting: (1) readily available support for offloaded training and inference \cite{ren2021zero, rajbhandari2021zero}, (2) clear trends towards more RAM and effective \textit{small} FMs \cite{applespecs2, tay2022unifyinglm}, and (3) many users own other capable devices beyond phones. 
Recently works demonstrate how to enable small and open-source models to meet the quality of 30x larger foundation models using in-context learning \cite{arora2022ama}.

\begin{table*}[t!]
    \begin{center}
    \normalsize
    \begin{tabular}{l|p{2.5cm}|p{4cm}|p{3.4cm}}
    \toprule
    Resource & {\centering  \systemname}
    
    (10B model) & {\centering \textsc{Federated Learning}}
    
    (100M model) &  Modern Phone  \\
    \midrule
    \multirow{1}{*}{Communication} & 
    \textbf{40 GB}, \textbf{1.5 hours} download & \textbf{40 GB, 13 hours} upload \& download, \textit{barring latency} 
    & 
    61 Mbps download \& 8 Mbps upload \\ 
    \midrule
    \multirow{1}{*}{Training} & 
    \textbf{None} & 
    \textbf{1e16 FLOPs}
    & 
    16 TFLOPs,
    4 GB RAM
    \\
    \midrule
    \multirow{1}{*}{Inference} & 
    \textbf{2e13 FLOPs}
    & 
    \textbf{1e11 FLOPs}
    &
    16 TFLOPs,
    4 GB RAM
    \\
    \midrule
    \multirow{1}{*}{Storage} & 
    \textbf{40 GB} on disk
    & 
    \textbf{400 MB} on disk 
    & 
    1 TB disk
    \\
    \bottomrule
    \end{tabular}
    \normalsize
    \caption{Cost comparison. Assumes 100 communication rounds, 32 batch size, 1000 steps, \& 512 sequence length for FL; 1024 max sequence length for the 10B model. Assumes full precision. Statistics for modern phones are from \citet{applespecs} and \cite{speedtest}.} 
    \label{tab:communication}
    \vspace{-3mm}
    \end{center}
\end{table*}

\textbf{Incentives} If the FM baseline is competitive even for a subset of the $N$ clients, the subset is disincentivized from participating. Not only do these clients avoid privacy leakage and communication costs of FL, but the fact that the zero-shot abilities succeed for these and not the other clients indicates there may be a distribution mismatch between the subsets of clients that make different decisions.

\section{Conclusion}
Amidst a recent focus on statistical notions of privacy, in light of in-context learning, perfect secrecy \textit{might} be possible for some sensitive tasks. While prior work initializes FL procedures from public pretrained models, we are the first to propose a complementary architecture based on ICL and no further training. There are several future challenges such as the fragility of prompting, out-of-domain degradation, and slow runtime of inference using large models. We release our code to help facilitate further study.

\section*{Acknowledgements}
We thank Sabri Eyuboglu, Neel Guha, Michael Zhang, Laurel Orr, Kawin Ethayarajh, Deepak Narayanan, Mayee Chen,  Maya Varma, Gary Cheng, Rohith Kuditipudi, Xuechen Li, Sidd Karamcheti, and Rishi Bommasani for their helpful feedback and discussions.
We gratefully acknowledge the support of DARPA under Nos. FA86501827865 (SDH) and FA86501827882 (ASED); NIH under No. U54EB020405 (Mobilize), NSF under Nos. CCF1763315 (Beyond Sparsity), CCF1563078 (Volume to Velocity), and 1937301 (RTML); ONR under No. N000141712266 (Unifying Weak Supervision); the Moore Foundation, NXP, Xilinx, LETI-CEA, Intel, IBM, Microsoft, NEC, Toshiba, TSMC, ARM, Hitachi, BASF, Accenture, Ericsson, Qualcomm, Analog Devices, the Okawa Foundation, American Family Insurance, Google Cloud, Swiss Re,
Brown Institute for Media Innovation, Department of Defense (DoD) through the National Defense Science and
Engineering Graduate Fellowship (NDSEG) Program,  Fannie and John Hertz Foundation,
National Science Foundation Graduate Research Fellowship Program, Texas Instruments, [Stanford Specific Fellowships], and members of the Stanford DAWN project: Teradata, Facebook, Google, Ant Financial, NEC, VMWare, and Infosys. The U.S. Government is authorized to reproduce and distribute reprints for Governmental purposes notwithstanding any copyright notation thereon. Any opinions, findings, and conclusions or recommendations expressed in this material are those of the authors and do not necessarily reflect the views, policies, or endorsements, either expressed or implied, of DARPA, NIH, ONR, or the U.S. Government.

\bibliographystyle{unsrtnat}
\bibliography{main}

\begin{thebibliography}{83}
\providecommand{\natexlab}[1]{#1}
\providecommand{\url}[1]{\texttt{#1}}
\expandafter\ifx\csname urlstyle\endcsname\relax
  \providecommand{\doi}[1]{doi: #1}\else
  \providecommand{\doi}{doi: \begingroup \urlstyle{rm}\Url}\fi

\bibitem[Bommasani et~al.(2021)Bommasani, Hudson, Adeli, Altman, Arora, von
  Arx, Bernstein, Bohg, Bosselut, Brunskill, and et~al.]{bommasani2021fm}
Rishi Bommasani, Drew~A. Hudson, E.~Adeli, Russ Altman, Simran Arora, S.~von
  Arx, Michael~S. Bernstein, Jeanette Bohg, A.~Bosselut, Emma Brunskill, and
  et~al.
\newblock On the opportunities and risks of foundation models.
\newblock \emph{arXiv:2108.07258}, 2021.

\bibitem[Wei et~al.(2022{\natexlab{a}})Wei, Tay, Bommasani, Raffel, Zoph,
  Borgeaud, Yogatama, Bosma, Zhou, Metzler, Chi, Hashimoto, Vinyals, Liang,
  Dean, and Fedus]{wei2022emergent}
Jason Wei, Yi~Tay, Rishi Bommasani, Colin Raffel, Barret Zoph, Sebastian
  Borgeaud, Dani Yogatama, Maarten Bosma, Denny Zhou, Donald Metzler, Ed~H.
  Chi, Tatsunori Hashimoto, Oriol Vinyals, Percy Liang, Jeff Dean, and William
  Fedus.
\newblock Emergent abilities of large language models, 2022{\natexlab{a}}.
\newblock URL \url{https://arxiv.org/abs/2206.07682}.

\bibitem[Shannon(1949)]{shannon1949secrecy}
Claude~E. Shannon.
\newblock Communication theory of secrecy systems.
\newblock 1949.
\newblock URL
  \url{https://pages.cs.wisc.edu/~rist/642-spring-2014/shannon-secrecy.pdf}.

\bibitem[Dodge et~al.(2020)Dodge, Ilharco, Schwartz, Farhadi, Hajishirzi, and
  Smith]{dodge2020smallfinetune}
Jesse Dodge, Gabriel Ilharco, Roy Schwartz, Ali Farhadi, Hannaneh Hajishirzi,
  and Noah Smith.
\newblock Fine-tuning pretrained language models: Weight initializations, data
  orders, and early stopping.
\newblock In \emph{arXiv:2002.06305}, 2020.
\newblock URL \url{https://arxiv.org/pdf/2002.06305.pdf}.

\bibitem[Shokri and Shmatikov(2015)]{shokri2015privacy}
Reza Shokri and Vitaly Shmatikov.
\newblock Privacy-preserving deep learning.
\newblock \emph{Proceedings of the 22nd ACM SIGSAC Conference on Computer and
  Communications Security (CCS)}, 2015.
\newblock URL \url{https://doi.org/10.1145/2810103.2813687}.

\bibitem[McMahan et~al.(2016)McMahan, Moore, Ramage, Hampson, and
  y~Arcas]{mcmahan2016fl}
H.~Brendan McMahan, Eider Moore, Daniel Ramage, Seth Hampson, and
  Blaise~Agüera y~Arcas.
\newblock Communication-efficient learning of deep networks from decentralized
  data.
\newblock \emph{Proceedings of the 20th International Conference on Artificial
  Intelligence and Statistics (AISTATS)}, 2016.

\bibitem[Wu et~al.(2022{\natexlab{a}})Wu, Li, Charles, Xiao, Liu, Xu, and
  Smith]{wu2022motley}
Shanshan Wu, Tian Li, Zachary Charles, Yu~Xiao, Ziyu Liu, Zheng Xu, and
  Virginia Smith.
\newblock Motley: Benchmarking heterogeneity and personalization in federated
  learning.
\newblock 2022{\natexlab{a}}.
\newblock URL \url{https://arxiv.org/abs/2206.09262}.

\bibitem[Shokri et~al.(2017)Shokri, Stronati, Song, and
  Shmatikov]{shokri2017meminference}
Reza Shokri, Marco Stronati, Congzheng Song, and Vitaly Shmatikov.
\newblock Membership inference attacks against machine learning models.
\newblock \emph{In the proceedings of the IEEE Symposium on Security and
  Privacy}, 2017.

\bibitem[Melis et~al.(2019)Melis, Song, Cristofaro, and
  Shmatikov]{melis2019flleakage}
Luca Melis, Congzheng Song, Emiliano~De Cristofaro, and Vitaly Shmatikov.
\newblock Exploiting unintended feature leakage in collaborative learning.
\newblock In \emph{Proceedings of 40th IEEE Symposium on Security and Privacy},
  2019.

\bibitem[Nasr et~al.(2019)Nasr, Shokri, and Houmansadr]{nasr2019meminference}
Milad Nasr, Reza Shokri, and Amir Houmansadr.
\newblock Comprehensive privacy analysis of deep learning: Passive and active
  white-box inference attacks against centralized and federated learning.
\newblock \emph{IEEE Symposium on Security and Privacy}, 2019.

\bibitem[Bagdasaryan et~al.(2020)Bagdasaryan, Veit, Hua, Estrin, and
  Shmatikov]{bagdasaryan2020backdoorfl}
Eugene Bagdasaryan, Andreas Veit, Yiqing Hua, Deborah Estrin, and Vitaly
  Shmatikov.
\newblock How to backdoor federated learning.
\newblock \emph{Proceedings of the Twenty Third International Conference on
  Artificial Intelligence and Statistics (PMLR)}, 2020.
\newblock URL \url{http://proceedings.mlr.press/v108/bagdasaryan20a.html}.

\bibitem[Dwork et~al.(2006)Dwork, McSherry, Nissim, and Smith]{dwork2006dp}
Cynthia Dwork, Frank McSherry, Kobbi Nissim, and Adam Smith.
\newblock Calibrating noise to sensitivity in private data analysis.
\newblock \emph{Theory of Cryptography Conference}, 2006.

\bibitem[Greenberg(2017)]{greenberg2017wired}
Andy Greenberg.
\newblock How one of apple's key privacy safeguards falls short, 2017.
\newblock URL
  \url{https://www.wired.com/story/apple-differential-privacy-shortcomings/}.

\bibitem[McMahan et~al.(2018)McMahan, Ramage, Talwar, and
  Zhang]{mcmahan2018dpfl}
H.~Brendan McMahan, Daniel Ramage, Kunal Talwar, and Li~Zhang.
\newblock Learning differentially private recurrent language models.
\newblock \emph{ICLR}, 2018.

\bibitem[Zhao et~al.(2018)Zhao, Li, Lai, Suda, Civin, and
  Chandra]{zhao2018noniid}
Yue Zhao, Meng Li, Liangzhen Lai, Naveen Suda, Damon Civin, and Vikas Chandra.
\newblock Federated learning with non-iid data.
\newblock \emph{arXiv:1806.00582}, 2018.

\bibitem[Arivazhagan et~al.(2019)Arivazhagan, Aggarwal, Singh, and
  Choudhary]{arivzhagan2019fedper}
Manoj~Ghuhan Arivazhagan, Vinay Aggarwal, Aaditya~Kumar Singh, and Sunav
  Choudhary.
\newblock Federated learning with personalization layers.
\newblock In \emph{arXiv:1912.00818v1}, 2019.
\newblock URL \url{https://arxiv.org/pdf/1912.00818.pdf}.

\bibitem[Li et~al.(2022{\natexlab{a}})Li, Hu, Beirami, and Smith]{li2021ditto}
Tian Li, Shengyuan Hu, Ahmad Beirami, and Virginia Smith.
\newblock Ditto: Fair and robust federated learning through personalization.
\newblock In \emph{ICML 2021}, 2022{\natexlab{a}}.
\newblock URL \url{https://arxiv.org/abs/2012.04221}.

\bibitem[Yin et~al.(2018)Yin, Chen, Ramchandran, and
  Bartlett]{yin2018byzantine}
Dong Yin, Yudong Chen, Kannan Ramchandran, and Peter Bartlett.
\newblock Byzantine-robust distributed learning: Towards optimal statistical
  rates.
\newblock In \emph{ICML}, 2018.

\bibitem[Konečný et~al.(2016)Konečný, McMahan, Yu, Richtárik, Suresh, and
  Bacon]{knecn2016fl}
Jakub Konečný, H.~Brendan McMahan, Felix~X. Yu, Peter Richtárik,
  Ananda~Theertha Suresh, and Dave Bacon.
\newblock Federated learning: Strategies for improving communication
  efficiency.
\newblock In \emph{arXiv:1610.05492}, 2016.

\bibitem[Zhan et~al.(2021)Zhan, Zhang, Hong, Wu, Li, and
  Guo]{zhan2021incentives}
Yufeng Zhan, Jie Zhang, Zicong Hong, Leijie Wu, Peng Li, and Song Guo.
\newblock A survey of incentive mechanism design for federated learning.
\newblock \emph{IEEE Transactions on Emerging Topics in Computing}, 2021.
\newblock URL
  \url{https://ieeexplore.ieee.org/stamp/stamp.jsp?arnumber=9369019}.

\bibitem[Brown et~al.(2020)Brown, Mann, Ryder, Subbiah, Kaplan, Dhariwal,
  Neelakantan, Shyam, Sastry, Askell, Agarwal, Herbert-Voss, Krueger, Henighan,
  Child, Ramesh, Ziegler, Wu, Winter, Hesse, Chen, Sigler, Litwin, Gray, Chess,
  Clark, Berner, McCandlish, Radford, Sutskever, and Amodei]{brown2020language}
Tom~B. Brown, Benjamin Mann, Nick Ryder, Melanie Subbiah, Jared Kaplan,
  Prafulla Dhariwal, Arvind Neelakantan, Pranav Shyam, Girish Sastry, Amanda
  Askell, Sandhini Agarwal, Ariel Herbert-Voss, Gretchen Krueger, Tom Henighan,
  Rewon Child, Aditya Ramesh, Daniel~M. Ziegler, Jeffrey Wu, Clemens Winter,
  Christopher Hesse, Mark Chen, Eric Sigler, Mateusz Litwin, Scott Gray,
  Benjamin Chess, Jack Clark, Christopher Berner, Sam McCandlish, Alec Radford,
  Ilya Sutskever, and Dario Amodei.
\newblock Language models are few-shot learners.
\newblock In \emph{arXiv:2005.14165}, 2020.

\bibitem[Radford et~al.(2021)Radford, Ki, Hallacy, Ramesh, Go, Agarwal, Sastry,
  Askell, Mishkin, Clark, Krueger, and Sutskever]{radford2021clip}
Alec Radford, Jong~Wook Ki, Chris Hallacy, Aditya Ramesh, Gabriel Go, Sandhini
  Agarwal, Girish Sastry, Amanda Askell, Pamela Mishkin, Jack Clark, Gretchen
  Krueger, and Ilya Sutskever.
\newblock Learning transferable visual models from natural language
  supervision.
\newblock \emph{arXiv:2103.00020}, 2021.

\bibitem[Chowdhery et~al.(2022)Chowdhery, Narang, Devlin, Bosma, Mishra,
  Roberts, Barham, Chung, Sutton, Gehrmann, et~al.]{chowdhery2022palm}
Aakanksha Chowdhery, Sharan Narang, Jacob Devlin, Maarten Bosma, Gaurav Mishra,
  Adam Roberts, Paul Barham, Hyung~Won Chung, Charles Sutton, Sebastian
  Gehrmann, et~al.
\newblock Palm: Scaling language modeling with pathways.
\newblock \emph{arXiv preprint arXiv:2204.02311}, 2022.

\bibitem[Liu et~al.(2022)Liu, Liu, Lu, Welleck, West, Bras, Choi, and
  Hajishirzi]{liu2022commonsense}
Jiacheng Liu, Alisa Liu, Ximing Lu, Sean Welleck, Peter West, Ronan~Le Bras,
  Yejin Choi, and Hannaneh Hajishirzi.
\newblock Generated knowledge prompting for commonsense reasoning.
\newblock 2022.
\newblock URL \url{https://liujch1998.github.io/assets/papers/GKP_v8.pdf}.

\bibitem[Liu et~al.(2021)Liu, Yuan, Fu, Jiang, Hayashi, and
  Neubig]{liu2021promptsurvey}
Pengfei Liu, Weizhe Yuan, Jinlan Fu, Zhengbao Jiang, Hiroaki Hayashi, and
  Graham Neubig.
\newblock Pre-train, prompt, and predict: A systematic survey of prompting
  methods in natural language processing.
\newblock In \emph{arXiv:2107.13586}, 2021.

\bibitem[Bell and LaPadula(1976)]{bell2976blm}
D.~E. Bell and L.~J. LaPadula.
\newblock Secure computer system: Unified exposition and multics
  interpretation.
\newblock \emph{The MITRE Corporation}, 1976.

\bibitem[Tian et~al.(2022)Tian, Wan, Lyu, Yao, Jin, and Sun]{tian2022fedbert}
Yuanyishu Tian, Yao Wan, Lingjuan Lyu, Dezhong Yao, Hai Jin, and Lichao Sun.
\newblock Fedbert: When federated learning meets pre-training.
\newblock In \emph{ACM Transactions on Intelligent Systems and Technology},
  2022.

\bibitem[Li et~al.(2022{\natexlab{b}})Li, Tramer, Liang, and
  Hashimoto]{li2022dpllm}
Xuechen Li, Florian Tramer, Percy Liang, and Tatsunori Hashimoto.
\newblock Large language models can be strong differentially private learners.
\newblock In \emph{ICLR 2022}, 2022{\natexlab{b}}.
\newblock URL \url{https://openreview.net/pdf?id=bVuP3ltATMz}.

\bibitem[Caldas et~al.(2019)Caldas, Meher, Duddu, Wu, Li, Konečný, McMahan,
  Smith, , and Talwalkar]{caldas2019leaf}
Sebastian Caldas, Sai Meher, Karthik Duddu, Peter Wu, Tian Li, Jakub Konečný,
  H.~Brendan McMahan, Virginia Smith, , and Ameet Talwalkar.
\newblock Leaf: A benchmark for federated settings.
\newblock In \emph{Workshop on Federated Learning for Data Privacy and
  Confidentiality}, 2019.

\bibitem[Miklau and Suciu(2004)]{miklau2004secrecy}
Gerome Miklau and Dan Suciu.
\newblock A formal analysis of information disclosure in data exchange.
\newblock \emph{SIGMOD}, 2004.
\newblock URL \url{https://homes.cs.washington.edu/~suciu/base-sigmod2004.pdf}.

\bibitem[Xu et~al.(2018)Xu, Qian, Mei, Huang, and Liu]{xu2018deeptype}
Mengwei Xu, Feng Qian, Qiaozhu Mei, Kang Huang, and Xuanzhe Liu.
\newblock Deeptype: On-device deep learning for input personalization service
  with minimal privacy concern.
\newblock \emph{Proceedings of the ACM on Interactive, Mobile, Wearable and
  Ubiquitous Technologies}, 2018.

\bibitem[Mosbach et~al.(2021)Mosbach, Andriushchenko, and
  Klakow]{mosbach2021smallfinetune}
Marius Mosbach, Maksym Andriushchenko, and Dietrich Klakow.
\newblock On the stability of fine-tuning bert: Misconceptions, explanations,
  and strong baselines.
\newblock In \emph{ICLR 2021}, 2021.
\newblock URL \url{https://arxiv.org/pdf/2006.04884.pdf}.

\bibitem[Li et~al.(2019)Li, Sahu, Talwalkar, and Smith]{li2019flsurvey}
Tian Li, Anit~Kumar Sahu, Ameet Talwalkar, and Virginia Smith.
\newblock Federated learning: Challenges, methods, and future directions.
\newblock 2019.
\newblock URL \url{https://arxiv.org/pdf/1908.07873.pdf}.

\bibitem[Fallah et~al.(2020)Fallah, Mokhtari, and Ozdaglar]{fallah2020fedmaml}
Alireza Fallah, Aryan Mokhtari, and Asuman Ozdaglar.
\newblock Personalized federated learning with theoretical guarantees: A
  model-agnostic meta-learning approach.
\newblock In \emph{34th Conference on Neural Information Processing Systems
  (NeurIPS 2020)}, 2020.
\newblock URL
  \url{https://proceedings.neurips.cc/paper/2020/file/24389bfe4fe2eba8bf9aa9203a44cdad-Paper.pdf}.

\bibitem[Hilmkil et~al.(2021)Hilmkil, Callh, Barbieri, Sutfeld, Zec, and
  Mogren]{hikmil2021fedfinetune}
Agrin Hilmkil, Sebastian Callh, Matteo Barbieri, Leon~Rene Sutfeld, Edvin~Listo
  Zec, and Olof Mogren.
\newblock Scaling federated learning for fine-tuning of large language models.
\newblock In \emph{arXiv:2102.00875}, 2021.

\bibitem[Chen et~al.(2021)Chen, Zhang, Guo, Fan, , and Cheng]{chen2021fedmatch}
Jiangui Chen, Ruqing Zhang, Jiafeng Guo, Yixing Fan, , and Xueqi Cheng.
\newblock Fedmatch: Federated learning over heterogeneous question answering
  data.
\newblock In \emph{Proceedings of the 30th ACM International Conference on
  Information and Knowledge Management (CIKM ’21),}, 2021.
\newblock URL
  \url{https://arxiv.org/pdf/2108.05069.pdf#page=10&zoom=100,76,202}.

\bibitem[Bonawitz et~al.(2017)Bonawitz, Ivanov, Kreuter, Marcedone, McMahan,
  Patel, Ramage, Segal, , and Seth]{bonawitz2017smpc}
Keith Bonawitz, Vladimir Ivanov, Ben Kreuter, Antonio Marcedone, H.~Brendan
  McMahan, Sarvar Patel, Daniel Ramage, Aaron Segal, , and Karn Seth.
\newblock Practical secure aggregation for privacy-preserving machine learning.
\newblock \emph{Proceedings of the 2017 ACM SIGSAC Conference on Computer and
  Communications Security (CCS)}, 2017.

\bibitem[Fang and Qian(2021)]{fang2021smpc}
Haokun Fang and Quan Qian.
\newblock Privacy preserving machine learning with homomorphic encryption and
  federated learning.
\newblock \emph{Future Internet}, 2021.
\newblock \doi{https://doi.org/10.3390/fi13040094}.

\bibitem[Kerrigan et~al.(2020)Kerrigan, Slack, and Tuyls]{kerrigan2020dplm}
Gavin Kerrigan, Dylan Slack, and Jens Tuyls.
\newblock Differentially private language models benefit from public
  pre-training.
\newblock In \emph{arXiv:2009.05886v2}, 2020.
\newblock URL \url{https://arxiv.org/abs/2009.05886}.

\bibitem[Yu et~al.(2020)Yu, Naik, Backurs, Gopi, Inan, Kamath, Kulkarni, Lee,
  Manoel, Wutschitz, Yekhanin, and Zhang]{yu2022dpfl}
Da~Yu, Saurabh Naik, Arturs Backurs, Sivakanth Gopi, Huseyin~A. Inan, Gautam
  Kamath, Janardhan Kulkarni, Yin~Tat Lee, Andre Manoel, Lukas Wutschitz,
  Sergey Yekhanin, and Huishuai Zhang.
\newblock Differentially private fine-tuning of language models.
\newblock In \emph{arXiv:2110.06500v2}, 2020.

\bibitem[Zhao et~al.(2021)Zhao, Wallace, Feng, Klein, and
  Singh]{zhao2021calibrate}
Tony~Z. Zhao, 1~Eric Wallace, Shi Feng, Dan Klein, and Sameer Singh.
\newblock Calibrate before use: Improving few-shot performance of language
  models.
\newblock In \emph{arXiv:2102.09690v2}, 2021.
\newblock URL \url{https://arxiv.org/pdf/2102.09690.pdf}.

\bibitem[Mishra et~al.(2021)Mishra, Khashabi, Baral, Choi, and
  Hajishirzi]{mishra2021reframing}
Swaroop Mishra, Daniel Khashabi, Chitta Baral, Yejin Choi, and Hannaneh
  Hajishirzi.
\newblock Reframing instructional prompts to gptk's language.
\newblock \emph{arXiv preprint arXiv:2109.07830}, 2021.

\bibitem[Wu et~al.(2022{\natexlab{b}})Wu, Terry, and Cai]{wu2022chains}
Tongshuang Wu, Michael Terry, and Carrie~J. Cai.
\newblock Ai chains: Transparent and controllable human-ai interaction by
  chaining large language model prompts.
\newblock In \emph{arXiv:2110.01691}, 2022{\natexlab{b}}.
\newblock URL \url{https://arxiv.org/pdf/2110.01691.pdf}.

\bibitem[Wei et~al.(2022{\natexlab{b}})Wei, Wang, Schuurmans, Bosma, Ichter,
  Xia, Chi, Le, and Zhou]{wei2022cot}
Jason Wei, Xuezhi Wang, Dale Schuurmans, Maarten Bosma, Brian Ichter, Fei Xia,
  Ed~H. Chi, Quoc~V. Le, and Denny Zhou.
\newblock Chain of thought prompting elicits reasoning in large language
  models.
\newblock \emph{arXiv:2201.11903v4}, 2022{\natexlab{b}}.

\bibitem[Hu et~al.(2006)Hu, Ferraiolo, and Kuhn]{hu2006nist}
Vincent~C. Hu, David~F. Ferraiolo, and D.~Rick Kuhn.
\newblock Assessment of access control systems.
\newblock \emph{NIST}, 2006.
\newblock URL \url{https://nvlpubs.nist.gov/nistpubs/Legacy/IR/nistir7316.pdf}.

\bibitem[He et~al.(2020)He, Li, So, Zhang, Wang, Wang, Vepakomma, Singh, Qiu,
  Shen, Zhao, Kang, Liu, Raskar, Yang, Annavaram, and
  Avestimehr]{chaoyanghe2020fedml}
Chaoyang He, Songze Li, Jinhyun So, Mi~Zhang, Hongyi Wang, Xiaoyang Wang,
  Praneeth Vepakomma, Abhishek Singh, Hang Qiu, Li~Shen, Peilin Zhao, Yan Kang,
  Yang Liu, Ramesh Raskar, Qiang Yang, Murali Annavaram, and Salman Avestimehr.
\newblock Fedml: A research library and benchmark for federated machine
  learning.
\newblock \emph{arXiv preprint arXiv:2007.13518}, 2020.

\bibitem[Lai et~al.(2021)Lai, Dai, Zhu, Madhyastha, and
  Chowdhury]{fedscale-arxiv}
Fan Lai, Yinwei Dai, Xiangfeng Zhu, Harsha~V. Madhyastha, and Mosharaf
  Chowdhury.
\newblock Fedscale: Benchmarking model and system performance of federated
  learning at scale.
\newblock In \emph{arXiv:2105.11367}, 2021.

\bibitem[Lin et~al.(2021)Lin, He, Zeng, Wang, Huang, Soltanolkotabi, Ren, and
  Avestimehr]{fednlp2021}
Bill~Yuchen Lin, Chaoyang He, ZiHang Zeng, Hulin Wang, Yufen Huang,
  M.~Soltanolkotabi, Xiang Ren, and S.~Avestimehr.
\newblock Fednlp: A research platform for federated learning in natural
  language processing.
\newblock In \emph{arXiv cs.CL 2104.08815}, 2021.
\newblock URL \url{https://arxiv.org/abs/2104.08815}.

\bibitem[Go et~al.(2009)Go, Bhayani, , and Huang]{go2009sent140}
Alec Go, Richa Bhayani, , and Lei Huang.
\newblock Twitter sentiment classification using distant supervision.
\newblock 2009.

\bibitem[Lang(1995)]{lang1995news20}
Ken Lang.
\newblock Newsweeder: Learning to filter netnews.
\newblock In \emph{In Proceedings of ICML}, 1995.

\bibitem[Liu et~al.(2015)Liu, Luo, Wang, , and Tang]{liu2015celeba}
Ziwei Liu, Ping Luo, Xiaogang Wang, , and Xiaoou Tang.
\newblock Deep learning face attributes in the wild.
\newblock 2015.

\bibitem[Krizhevsky(2009)]{krizhevsky2009cifar10}
Alex Krizhevsky.
\newblock Learning multiple layers of features from tiny images, 2009.
\newblock URL \url{https://www.cs.toronto.edu/~kriz/cifar.html}.

\bibitem[Cohen et~al.(2017)Cohen, Afshar, Tapson, , and van
  Schaik]{cohen2017emnist}
Gregory Cohen, Saeed Afshar, Jonathan Tapson, , and André van Schaik.
\newblock Emnist: an extension of mnist to handwritten letters.
\newblock 2017.

\bibitem[Fisch et~al.(2019)Fisch, Talmor, Jia, Seo, Choi, and
  Chen]{fisch2019mrqa}
Adam Fisch, Alon Talmor, Robin Jia, Minjoon Seo, Eunsol Choi, and Danqi Chen.
\newblock Mrqa 2019 shared task: Evaluating generalization in reading
  comprehension.
\newblock 2019.

\bibitem[Sanh et~al.(2022)Sanh, Webson, Raffel, Bach, Sutawika, Alyafeai,
  Chaffin, Stiegler, Scao, Raja, and et. al.]{sanh2022t0}
Victor Sanh, Albert Webson, Colin Raffel, Stephan~H. Bach, Lintang Sutawika,
  Zaid Alyafeai, Antoine Chaffin, Arnaud Stiegler, Teven~Le Scao, Arun Raja,
  and et. al.
\newblock Multitask prompted training enables zero-shot task generalization.
\newblock In \emph{ICLR 2022}, 2022.
\newblock URL \url{https://arxiv.org/pdf/2110.08207.pdf}.

\bibitem[Reimers and Gurevych(2019)]{reimers-2019-sentence-bert}
Nils Reimers and Iryna Gurevych.
\newblock Sentence-bert: Sentence embeddings using siamese bert-networks.
\newblock In \emph{Proceedings of the 2019 Conference on Empirical Methods in
  Natural Language Processing}. Association for Computational Linguistics, 11
  2019.
\newblock URL \url{https://arxiv.org/abs/1908.10084}.

\bibitem[Song et~al.(2020)Song, Tan, Qin, Lu, and Liu]{song2020mpnet}
Kaitao Song, Xu~Tan, Tao Qin, Jianfeng Lu, and Tie-Yan Liu.
\newblock Mpnet: Masked and permuted pre-training for language understanding.
\newblock In \emph{arXiv:2004.09297}, 2020.
\newblock URL \url{https://arxiv.org/abs/2004.09297}.

\bibitem[Zellers et~al.(2019)Zellers, Holtzman, Rashkin, Bisk, Farhadi,
  Roesner, and Choi]{zellers2019grover}
Rowan Zellers, Ari Holtzman, Hannah Rashkin, Yonatan Bisk, Ali Farhadi,
  Franziska Roesner, and Yejin Choi.
\newblock Defending against neural fake news.
\newblock In \emph{Advances in Neural Information Processing Systems 32}, 2019.

\bibitem[Michieli and Ozay(2021)]{michieli2021sent140baseline}
Umberto Michieli and Mete Ozay.
\newblock Are all users treated fairly in federated learning systems?
\newblock In \emph{CVPR}, 2021.
\newblock URL
  \url{https://openaccess.thecvf.com/content/CVPR2021W/RCV/papers/Michieli_Are_All_Users_Treated_Fairly_in_Federated_Learning_Systems_CVPRW_2021_paper.pdf}.

\bibitem[Qu et~al.(2022)Qu, Zhou, Liang, Xia, Wang, Adeli, Fei-Fei, and
  Rubin]{qu2022vitfl}
Liangqiong Qu, Yuyin Zhou, Paul~Pu Liang, Yingda Xia, Feifei Wang, Ehsan Adeli,
  Li~Fei-Fei, and Daniel Rubin.
\newblock Rethinking architecture design for tackling data heterogeneity in
  federated learning.
\newblock In \emph{CVPR 2022}, 2022.
\newblock URL \url{https://arxiv.org/abs/2106.06047}.

\bibitem[Rothchild et~al.(2020)Rothchild, Panda, Ullah, Ivkin, Stoica,
  Braverman, Gonzalez, and Arora]{rothchild2020fetchsgd}
Daniel Rothchild, Ashwinee Panda, Enayat Ullah, Nikita Ivkin, Ion Stoica,
  Vladimir Braverman, Joseph Gonzalez, and Raman Arora.
\newblock Fetchsgd: Communication-efficient federated learning with sketching.
\newblock In \emph{Proceedings of the 37th International Conference on Machine
  Learning (PMLR)}, 2020.
\newblock URL
  \url{http://proceedings.mlr.press/v119/rothchild20a/rothchild20a.pdf}.

\bibitem[Sun et~al.(2021)Sun, Li, and Wang]{sun2021decentralizedfl}
Taro Sun, Dongsheng Li, and Bao Wang.
\newblock Decentralized federated averaging.
\newblock In \emph{arXiv:2104.11375v1}, 2021.
\newblock URL \url{https://arxiv.org/abs/2012.04221}.

\bibitem[Bach et~al.(2022)Bach, Sanh, Yong, Webson, Raffel, Nayak, Sharma, Kim,
  Bari, Fevry, Alyafeai, Dey, Santilli, Sun, Ben-David, Xu, Chhablani, Wang,
  Fries, Al-shaibani, Sharma, Thakker, Almubarak, Tang, Tang, Jiang, and
  Rush]{bach2022promptsource}
Stephen~H. Bach, Victor Sanh, Zheng-Xin Yong, Albert Webson, Colin Raffel,
  Nihal~V. Nayak, Abheesht Sharma, Taewoon Kim, M~Saiful Bari, Thibault Fevry,
  Zaid Alyafeai, Manan Dey, Andrea Santilli, Zhiqing Sun, Srulik Ben-David,
  Canwen Xu, Gunjan Chhablani, Han Wang, Jason~Alan Fries, Maged~S.
  Al-shaibani, Shanya Sharma, Urmish Thakker, Khalid Almubarak, Xiangru Tang,
  Xiangru Tang, Mike Tian-Jian Jiang, and Alexander~M. Rush.
\newblock Promptsource: An integrated development environment and repository
  for natural language prompts, 2022.

\bibitem[Gao et~al.(2021)Gao, Biderman, Black, Golding, Hoppe, Foster, Phang,
  He, Thite, Nabeshima, Presser, and Leahy]{gao2021pile}
Leo Gao, Stella Biderman, Sid Black, Laurence Golding, Travis Hoppe, Charles
  Foster, Jason Phang, Horace He, Anish Thite, Noa Nabeshima, Shawn Presser,
  and Connor Leahy.
\newblock The pile: An 800gb dataset of diverse text for language modeling,
  2021.
\newblock URL \url{https://arxiv.org/abs/2101.00027}.

\bibitem[Delateur(2019)]{delateur2019backpage}
Monica~J. Delateur.
\newblock From craigslist to backpage.com: Conspiracy as a strategy to
  prosecute third-party websites for sex trafficking.
\newblock In \emph{Santa Clara Law Review}, pages 532--590, 2019.
\newblock URL
  \url{https://digitalcommons.law.scu.edu/cgi/viewcontent.cgi?article=2826&context=lawreview}.

\bibitem[doe(2016)]{doevbackpage2016}
Jane doe, et al., petitioners v. backpage.com llc, et al.
\newblock In \emph{In the Supreme Court of the United States}, 2016.

\bibitem[Ren et~al.(2021)Ren, Rajbhandari, Aminabadi, Ruwase, Yang, Zhang, Li,
  and He]{ren2021zero}
Jie Ren, Samyam Rajbhandari, Reza~Yazdani Aminabadi, Olatunji Ruwase, Shuangyan
  Yang, Minjia Zhang, Dong Li, and Yuxiong He.
\newblock Zero-offload: Democratizing billion-scale model training.
\newblock In \emph{arXiv:2101.06840}, 2021.
\newblock URL \url{https://arxiv.org/abs/2101.06840}.

\bibitem[Rajbhandari et~al.(2021)Rajbhandari, Ruwase, Rasley, Smith, and
  He]{rajbhandari2021zero}
Samyam Rajbhandari, Olatunji Ruwase, Jeff Rasley, Shaden Smith, and Yuxiong He.
\newblock Zero-infinity: Breaking the gpu memory wall for extreme scale deep
  learning.
\newblock In \emph{arXiv:2104.07857v1}, 2021.
\newblock URL \url{https://arxiv.org/pdf/2104.07857.pdf}.

\bibitem[app(2022)]{applespecs2}
Blake's ios device specifications grid, April 2022.
\newblock URL \url{https://blakespot.com/ios_device_specifications_grid.html}.

\bibitem[Tay et~al.(2022)Tay, Dehghani, Tran, Garcia, Bahri, Schuster, Zheng,
  Houlsby, and Metzler]{tay2022unifyinglm}
Yi~Tay, Mostafa Dehghani, Vinh~Q. Tran, Xavier Garcia, Dara Bahri, Tal
  Schuster, Huaixiu~Steven Zheng, Neil Houlsby, and Donald Metzler.
\newblock Unifying language learning paradigms.
\newblock \emph{arXiv:2205.05131v1}, 2022.
\newblock URL \url{https://arxiv.org/pdf/2205.05131.pdf}.

\bibitem[Arora et~al.(2022)Arora, Narayan, Chen, Orr, Guha, Bhatia, Chami,
  Sala, and R\'e]{arora2022ama}
Simran Arora, Avanika Narayan, Mayee~F. Chen, Laurel Orr, Neel Guha, Kush
  Bhatia, Ines Chami, Frederic Sala, and Christopher R\'e.
\newblock Ask me anything: A simple strategy for prompting language models.
\newblock \emph{arXiv:2210.02441}, 2022.

\bibitem[Freedman(2021)]{applespecs}
Andrew~E. Freedman.
\newblock Apple a15 bionic powers iphone 13 and ipad mini, September 2021.
\newblock URL
  \url{https://www.tomshardware.com/news/ipad-iphone-13-a15-bionic}.

\bibitem[spe(2022)]{speedtest}
United states' mobile and fixed broadband internet speeds, 2022.
\newblock URL \url{https://www.speedtest.net/global-index/united-states}.

\bibitem[PyTorch(2022)]{pytorchmac}
PyTorch.
\newblock Introducing accelerated pytorch training on mac, May 2022.
\newblock URL
  \url{https://pytorch.org/blog/introducing-accelerated-pytorch-training-on-mac/}.

\bibitem[Peddie and Dow(2022)]{peddieGPU}
Jon Peddie and Robert Dow.
\newblock Q1’22 saw a decline in gpu and pc shipments quarter-to-quarter, May
  2022.
\newblock URL
  \url{https://www.jonpeddie.com/press-releases/q122-saw-a-decline-in-gpu-and-pc-shipments-quarter-to-quarter#:~:text=The\%20GPU's\%20overall\%20attach\%20rate,\%25\%20year\%2Dto\%2Dyear.}

\bibitem[fol(2022)]{foldingathome}
Os statistics., July 2022.
\newblock URL \url{https://stats.foldingathome.org/os}.

\bibitem[Diskin et~al.(2021)Diskin, Bukhtiyarov, Ryabinin, Saulnier, Sinitsin,
  Popov, Pyrkin, Kashirin, Borzunov, and Albert Villanova~del
  Moral]{diskin2021distributed}
Michael Diskin, Alexey Bukhtiyarov, Max Ryabinin, Lucile Saulnier, Anton
  Sinitsin, Dmitry Popov, Dmitry~V Pyrkin, Maxim Kashirin, Alexander Borzunov,
  and et~al. Albert Villanova~del Moral.
\newblock Distributed deep learning in open collaborations.
\newblock In \emph{arXiv:2106.10207v2}, 2021.

\bibitem[Ouyang et~al.(2022)Ouyang, Wu, Jiang, Almeida, Wainwright, Mishkin,
  Zhang, Agarwal, Slama, Ray, Schulman, Hilton, Kelton, Miller, and
  et~al.]{ouyang2022instruct}
Long Ouyang, Jeff Wu, Xu~Jiang, Diogo Almeida, Carroll~L. Wainwright, Pamela
  Mishkin, Chong Zhang, Sandhini Agarwal, Katarina Slama, Alex Ray, John
  Schulman, Jacob Hilton, Fraser Kelton, Luke Miller, and et~al.
\newblock Training language models to follow instructions with human feedback.
\newblock In \emph{arXiv:2203.02155v1}, 2022.

\bibitem[Zhang et~al.(2022)Zhang, Roller, Goyal, Artetxe, Chen, Chen, Dewan,
  Diab, Li, Lin, and et. al.]{zhang2022opt}
Susan Zhang, Stephen Roller, Naman Goyal, Mikel Artetxe, Moya Chen, Shuohui
  Chen, Christopher Dewan, Mona Diab, Xian Li, Xi~Victoria Lin, and et. al.
\newblock Opt: Open pre-trained transformer language models.
\newblock In \emph{arXiv:2205.01068}, 2022.
\newblock URL \url{https://arxiv.org/abs/2205.01068}.

\bibitem[OpenAI(2021)]{openai_2021}
OpenAI.
\newblock Openai api, Nov 2021.
\newblock URL \url{https://openai.com/api/}.

\bibitem[Wei et~al.(2019)Wei, Li, Ding, Ma, Yang, Farhad, Jin, Quek, and
  Poor]{wei2019dpfl}
Kang Wei, Jun Li, Ming Ding, Chuan Ma, Howard~H. Yang, Farokhi Farhad, Shi Jin,
  Tony Q.~S. Quek, and H.~Vincent Poor.
\newblock Federated learning with differential privacy: Algorithms and
  performance analysis.
\newblock \emph{arXiv:1911.00222}, 2019.
\newblock URL \url{https://arxiv.org/abs/1911.00222v2}.

\bibitem[Bagdasaryan et~al.(2019)Bagdasaryan, Veit, Hua, Estrin, and
  Shmatikov]{bagdasaryan2019flpoison}
Eugene Bagdasaryan, Andreas Veit, Yiqing Hua, Deborah Estrin, and Vitaly
  Shmatikov.
\newblock How to backdoor federated learning.
\newblock \emph{arXiv:1807.00459v3}, 2019.
\newblock URL \url{https://arxiv.org/pdf/1807.00459.pdf}.

\bibitem[Xu and Lyu(2020)]{xu2020fairfl}
Xinyi Xu and Lingjuan Lyu.
\newblock Towards building a robust and fair federated learning system.
\newblock \emph{arXiv:2011.10464v1}, 2020.
\newblock URL \url{https://arxiv.org/pdf/2011.10464v1.pdf}.

\end{thebibliography}


\newpage
\vspace{-3mm}
\section{Appendix: Experimental Details}
\label{sec:appendix}
\subsection{Benchmark Protocols}

\begin{table*}[t!]
        \begin{center}
    \normalsize
    \begin{tabular}{lcc}
    \toprule
      \multirow{1}{*}{Privacy Benchmark}  &  Train Examples per User  & Total Test Set Size \\
    \midrule
    Reddit (Non-IID) \cite{caldas2019leaf}  & 34.1  &  25.5k \\
    MRQA  (Non-IID)  \cite{fisch2019mrqa}  & 501.5 &  3.0k \\
    Sent140 (Non-IID) \cite{go2009sent140} & 2.4   &  286.6k \\
    20News (IID)  \cite{lang1995news20} & 113.1 &  7.5k \\
    CelebA (IID) \cite{liu2015celeba} & 21.4  & 11.1k\\
    CIFAR-10 (IID) \cite{krizhevsky2009cifar10} & Varied & 10.0k \\
    F-EMNIST (IID) \cite{cohen2017emnist} & 226.8 &  81.7k \\
    \bottomrule
    \end{tabular}
    \normalsize
    \caption{We show statistics for each benchmark and whether each has IID or Non-IID data per user.}. 
    \label{tab:benchmarksizes}
    \end{center}
\end{table*}

We release our code for reproducability and future work, including task and class descriptions we use, prompt formats, and task scoring functions. The benchmarks were originally proposed in \citet{caldas2019leaf, fednlp2021} and are openly-accessible.\footnote{\url{https://github.com/FedML-AI/FedNLP}} \footnote{ \url{https://github.com/TalwalkarLab/leaf}} Benchmark size-statistics are shown in Table \ref{tab:benchmarksizes} and here we provide details of any dataset sampling or prompting choice.

\textbf{CelebA} This is an image classification benchmark and we evaluate using the CLIP-ViT32B model using the full test dataset. We encode the binary class descriptions and match the image embeddings to the closest class embedding.
\begin{lstlisting}
task_description = NA 
class_descriptions = ["frowning", "smiling"]
\end{lstlisting}

\textbf{CIFAR-10} This is an image classification benchmark and we evaluate using the CLIP-ViT32B model using the full test dataset. We encode the class descriptions and match the image embeddings to the closest class embedding.
\begin{lstlisting}
task_description = NA 
class_descriptions = ["airplane", "automobile", "bird", "cat", "deer", "dog", "frog", "horse", "ship", "truck"]
\end{lstlisting}

\textbf{F-EMNIST}  This is an image classification benchmark and we evaluate using the CLIP-ViT32B model, using the full test dataset. We encode the class descriptions and match the image embeddings to the closest class embedding. 
\textit{Note that the CLIP models are not case-sensitive.}
\begin{lstlisting}
task_description = NA 
class_descriptions = ["The picture is of the uppercase letter <<c>>", "The picture is of the lowercase letter <<c>>", "The picture is of the digit <<c>>"] where the classname c is inserted as <<c>>
\end{lstlisting}

\textbf{Sent140} This is a text classification benchmark and we evaluate using the GPT, T0, and bi-encoder model variants, using a constant random subsample of 2.5\% of clients (7.2k examples), due to cost restrictions. We include clients with $\geq 0$ labeled training examples. For the autoregressive models, we map the output to a class using the first generated token (scoring code is released). For the bi-encoder model, we encode class descriptions ``positive'' and ```negative'' and match the input embeddings to the closest class embedding. The fixed public prompt used in Figure \ref{fig:gptreddit} is shown below:

\begin{lstlisting}
task_description = "Is this text positive or negative? Text: " 
class_descriptions = ["positive", "negative"]
\end{lstlisting}

\textbf{20News} This is a text classification benchmark and we evaluate using the GPT, T0, and bi-encoder model variants, using the full test dataset. For the autoregressive models, we map the output to a class using the first generated token (scoring code is released). For the bi-encoder model, we directly encode class descriptions as follows and match the input embeddings to the closest class embedding.
\begin{lstlisting}
task_description = "Is the topic electronics, cryptography security, religion, graphics, ibm pc hardware, space, politics, mac hardware, motorcycles, politics middle east, sale, automobiles, baseball, medical, christianity, os ms-windows, politics guns, atheism, hockey, or windows x? " 
class_descriptions = ["electronics", "cryptography security", "religion", "graphics", "ibm pc hardware", "space", "politics", "mac hardware", "motorcycles", "politics middle east", "sale", "automobiles", "baseball", "medical", "christianity", "os ms-windows", "politics guns", "atheism", "hockey", "windows x"]
\end{lstlisting}

\textbf{Synthetic 20News} This is a text classification synthetic dataset constructed by subsampling 1k points from the 20News dataset that have one of the desired labels. We evaluate using GPT-6.7B and GPT-175B. We specifically create two synthetics, termed ``Coarse'' and ``Fine''. The Coarse details are as follows:
\begin{lstlisting}
task_description = "Is the topic politics, baseball, medical, or automobiles? " 
class_descriptions = ["politics", "baseball", "medical", "automobiles"]
\end{lstlisting}

The Fine details are as follows:
\begin{lstlisting}
task_description = "Is the topic politics, religion, politics guns, or politics middle east? " 
class_descriptions = ["politics", "religion", "politics guns", "politics middle east"]
\end{lstlisting}

\textbf{MRQA} This is a reading comprehension benchmark and we evaluate using GPT model, using the full test dataset. Given an input question and context, we prompt the model as follows:
\begin{lstlisting}
task_description = 
    "What is the answer span to the following question?

    Obama was born in Honolulu, Hawaii. After graduating from Columbia University in 1983, he worked as a community organizer in Chicago. In 1988, he enrolled in Harvard Law School, where he was the first black president of the Harvard Law Review. After graduating, he became a civil rights attorney and an academic, teaching constitutional law at the University of Chicago Law School from 1992 to 2004. Turning to elective politics, he represented the 13th district in the Illinois Senate from 1997 until 2004, when he ran for the U.S. Senate. Obama received national attention in 2004 with his March Senate primary win, his well-received July Democratic National Convention keynote address, and his landslide November election to the Senate.

    Question: When was Barack Obama born?
    Answer: 1983
    
    <<context>>
    
    Question: <<input>>
    Answer: "
    
class_descriptions = NA
\end{lstlisting}

\textbf{Reddit} This is a language modeling benchmark and we evaluate using the Grover and GPT variants, using  the full test dataset. However, for GPT-6.7B and GPT-175B we subsample 5\% of the clients.
\begin{lstlisting}
task_description = NA
class_descriptions = NA
\end{lstlisting}

\subsection{Experimental Details}
\paragraph{Inference-only strategies} 
For the results in Table \ref{tab:benchmarks} and Figure \ref{fig:sizes}, we use the models zero-shot. For autoregressive models, we simply prompt the model with the exact \textit{task descriptions} specified above. For similarity search, we use the exact \textit{class descriptions} specified above. For all results in Figure \ref{fig:gptreddit}, we use \textit{no task description} to focus on the effect of \textit{task demonstrations}. Note that to hold the number of task demonstrations per user constant across baselines, the ``No User Privacy'' baseline randomly selects the minimum of \{number of user's training examples, $k$\} for $k \in \{3, 5\}$ --- for example on Sent140, the average number of task demonstrations per user is 2.4, and the number of in-context examples available for the ``User Privacy'' baseline is the user's training data size.




\subsection{Examples of each few-sample adaptation technique}

Using a topic classification task as an example, the following demonstrate examples of (1) prompting with task descriptions or labeled examples, (2) zero-shot classification via nearest neighbors similarity search, and (3) FM tuning.

An example of prompting follows. The above prompt includes the task description and the below prompt includes task demonstrations. 
    \begin{lstlisting}
    // With Task Instructions (Zero Shot) 
    prompt = "
        Is the text about medicine, baseball, atheism, electronics, ..., or hockey?
        Text: I believe the Cubs have the best record ever in the MLB.
    "
    prediction = model.generate(prompt)
    
    //  With Task Demonstrations (Few Shot)
    prompt = "
        Is the text about medicine, baseball, atheism, electronics, ..., or hockey?
        
        Input: They detect the oscillator operating in the detector.
        Label: electronics
        
        Input: I believe the Cubs have the best record ever in the MLB.
        Label: 
    "
    prediction = model.generate(prompt)
    
    >> print(prediction)
    >> "baseball"
    \end{lstlisting}

An example of similarity search follows. 
    \begin{lstlisting}
    input = "I believe the Cubs have the best record ever in the MLB."
    class_descriptions = ["baseball", "hockey", "politics in the middle east", "electronics", "medicine", "automobiles", "atheism", "cryptography"]
    
    embedded_input = model.encode_query(input)
    embedded_descriptions =  model.encode_keys(class_descriptions)
    
    // predict class with highest embedding similarity score to the input
    prediction = max_similarity_class(encoded_input, encoded_descriptions)
    
    >> print(prediction)
    >> "baseball"
    
    \end{lstlisting}

\subsection{Additional Discussion of \systemname{} Deployment}

Here we discuss additional considerations for deploying \systemname{}, elaborating on the systems feasibility, FM availability, and FM prompting reliability challenges.

\paragraph{Systems Feasibility} In Section \ref{sec: systems}, we observe it would be highly inefficient to use large FMs and FL on modern mobile phones. However, we refer to the fact that several users and organizations also own devices beyond mobile phones. In particular, modern personal computers are capable of performing training and inference for several-hundred million parameter models \cite{pytorchmac}. Further, the personal computer and desktop GPU market shipped 96 million units in Q1 2022 \cite{peddieGPU}. The availability of GPUs by a large number of parties is evidenced by the increasing number of projects that take advantage of available and underutilized personal GPUs \cite{foldingathome, diskin2021distributed}. Large FMs can also be hosted on CPUs during inference. Meanwhile, hardware is improving at a rapid rate \cite{applespecs}.

Meanwhile, recent progress has showed that fine-tuning smaller FMs to follow prompt-instructions improves zero-shot generalization. For instance, finetuned models ranging from 1.3B to 11B outperform their 175B pretrained counterparts \cite{ouyang2022instruct}. Together trends point towards  improved accessibility to FM baselines over time, however deploying current billion-parameter FMs is challenging.

\paragraph{FM Availability} \systemname{} relies on the availability of an FM. FMs both trained over general web data \cite{zhang2022opt} and over domain-specific data (e.g., medical, legal) are increasingly available. These resources expand the scope in which \systemname{} may be successful, however FMs are not available for all domains and many current FMs are only available via API access, which does not protect data-privacy \cite{openai_2021}. We are optimistic about the increasing availability of open-source models \cite{zhang2022opt}.

\paragraph{FM Interfaces and Reliability} The next key deployment challenge for \systemname{} is to develop interfaces for users to express prompts and integrate personal data-sources. Current FMs have largely been used by ML practitioners and prompt engineering is a temperamental process, especially for more complex tasks of interest. Progress on human-computer interfaces for FMs is improving the user experience. For instance, \citet{wu2022chains} propose to decompose complex tasks into sub-steps, each of which is easier for users to prompt and debug. However, decomposition-based methods introduce additional inference-calls to the FM (i.e. a new call for each sub-step), which increases latency. These are interesting tradeoffs to study in future work.

\paragraph{Improving Quality} As shown in Section \ref{sec:experiments}, FM performance drastically underperforms FL on fine-grained classification tasks, using standard baselines. In Appendix \ref{sec:appendix_baselines}, we provide additional experiments using more advanced FM-prompting strategies.

Finally, we note that as users interact with models, their outward behaviors and decisions may change over time. This can occur under any privacy framework or non-private modes of operation. However, perfect secrecy is defined with respect the system providing the decisions and does not encompass all external user actions \cite{shannon1949secrecy}.

\subsection{Extended Analysis}
\label{sec:appendix_results}

\begin{figure*}
    \centering
    \includegraphics[width=0.49\linewidth]{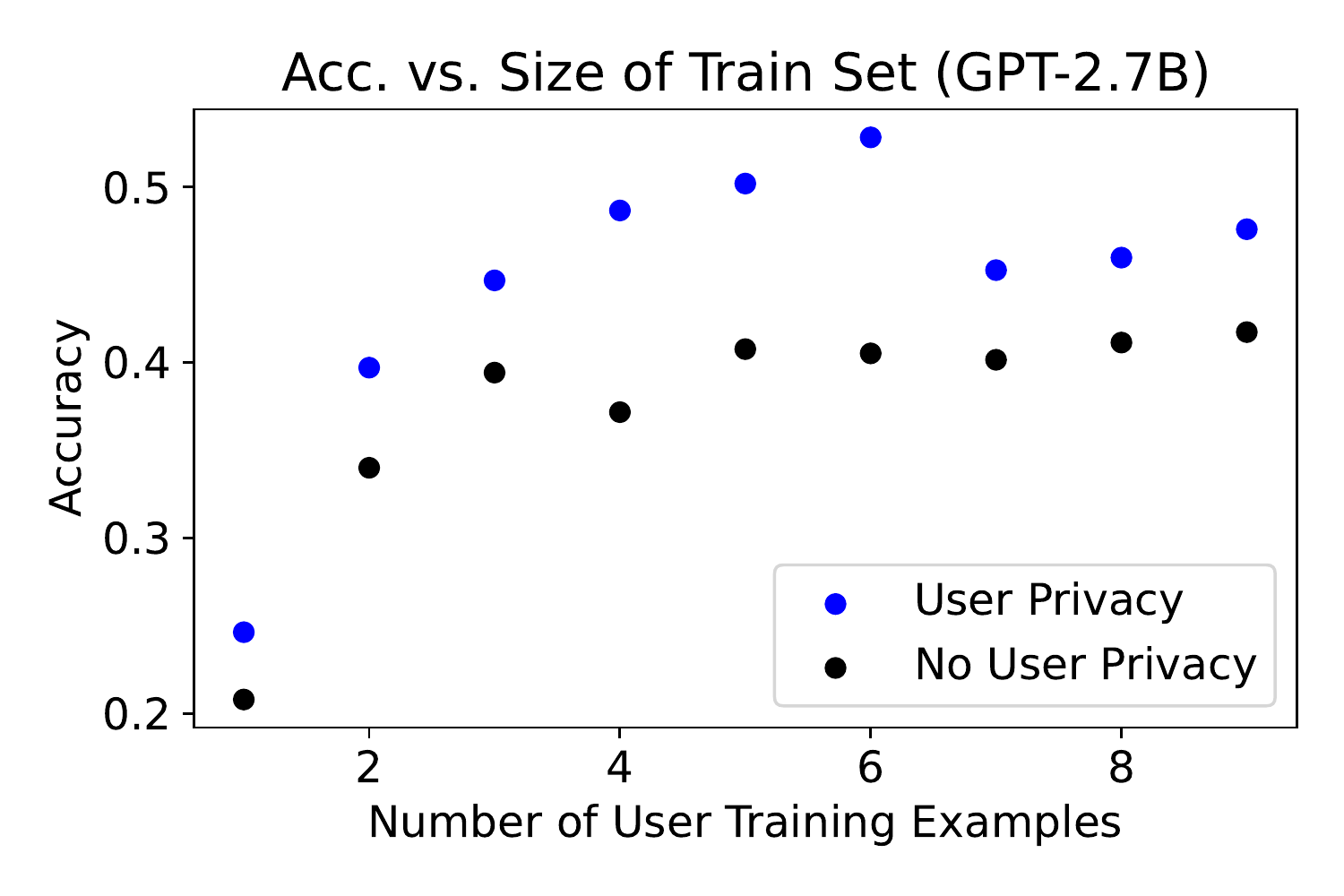}
    \includegraphics[width=0.49\linewidth]{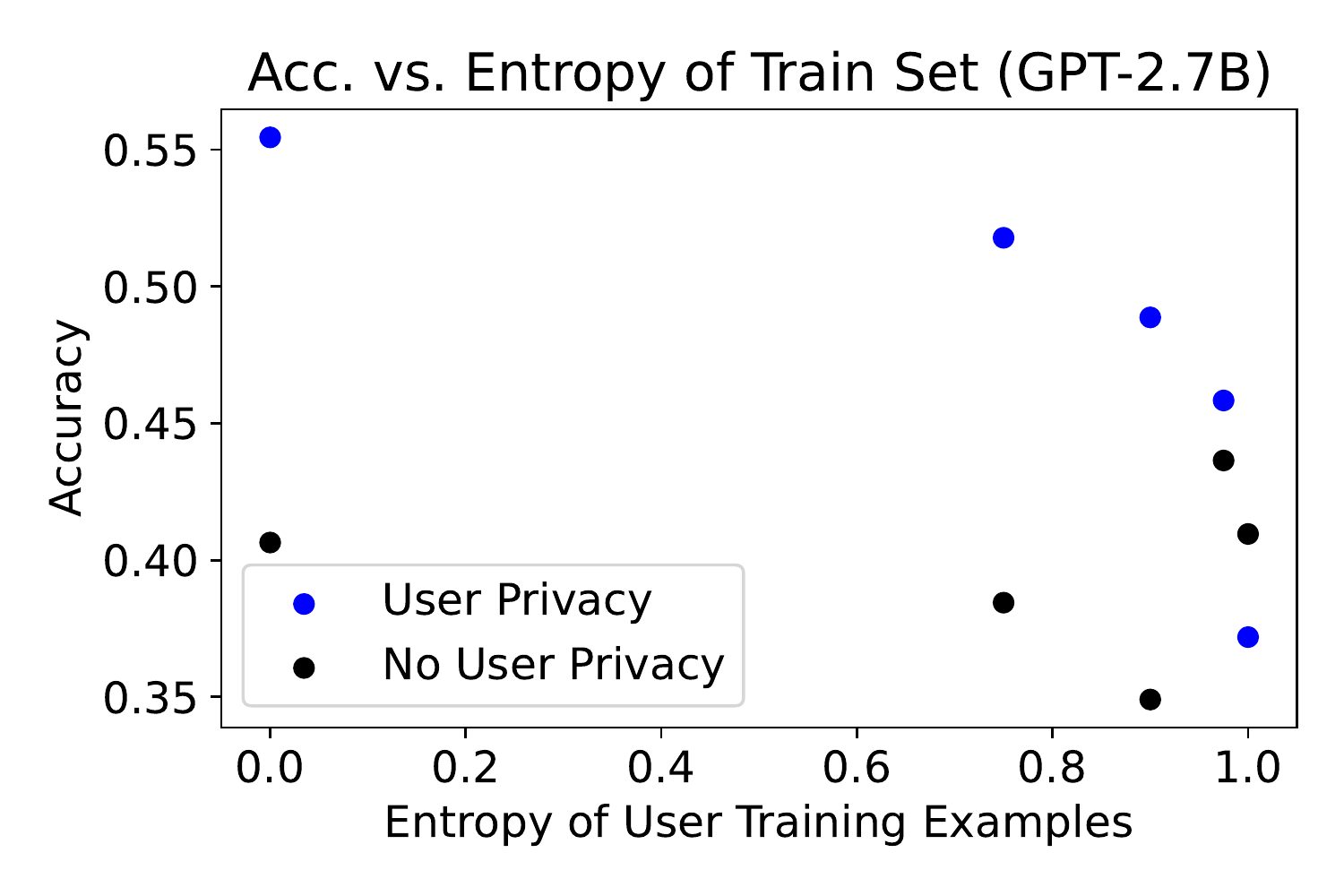}
    \includegraphics[width=0.49\linewidth]{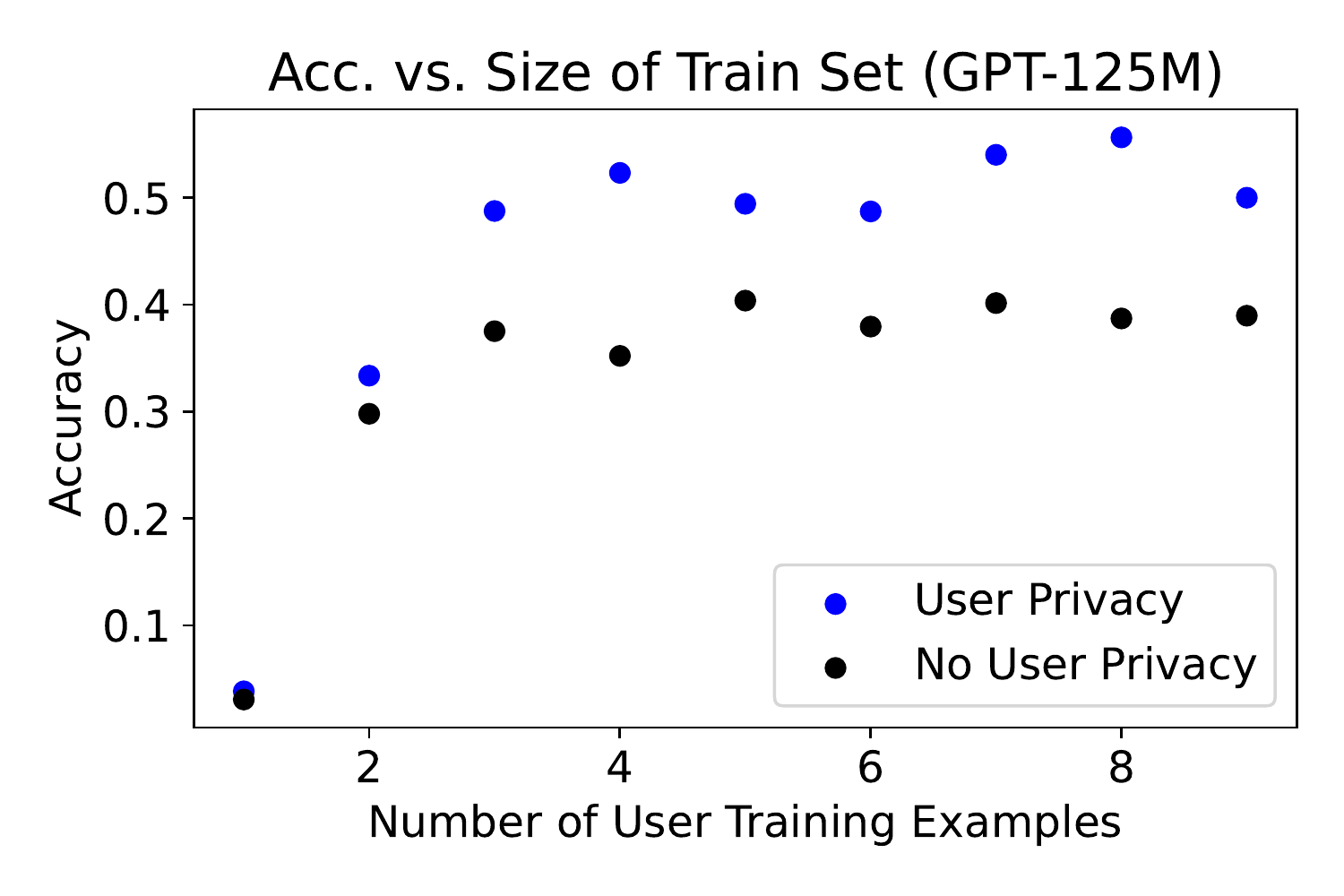}
    \includegraphics[width=0.49\linewidth]{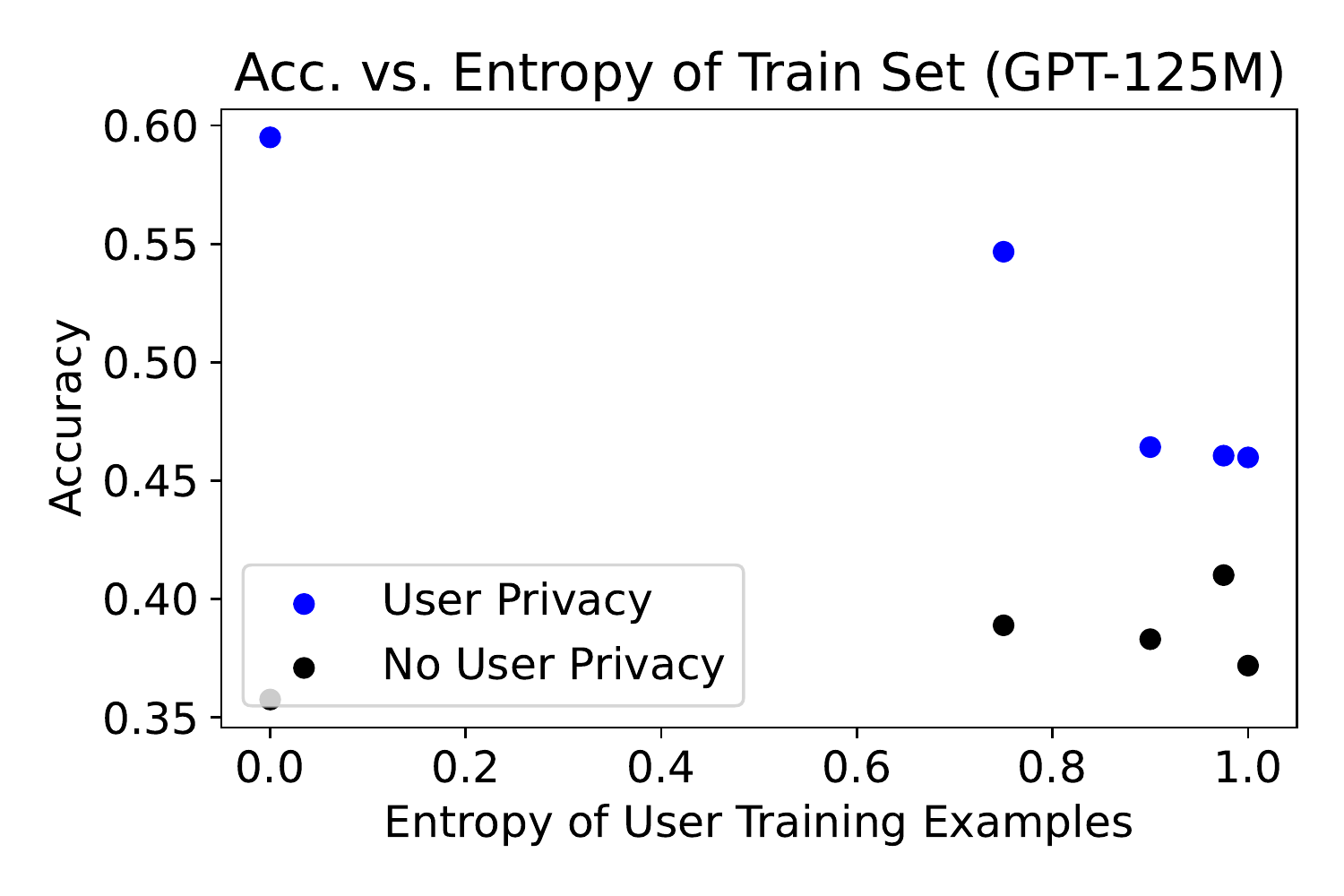}
    \caption[width=\linewidth]{Accuracy vs. user-level training data properties.}
    \label{fig:whyuserlevel}
\end{figure*}

\begin{table*}[t!]
        \begin{center}
    \normalsize
    \begin{tabular}{p{2cm}p{11cm}}
    \toprule
      \multirow{1}{*}{User}  &  Task Demonstrations \\
    \midrule
    User 1 \newline (Sent140, \newline Syntax) &  
    \textit{Example 1} @MYIDOLTOWN Oh my gosh!!!!!!!!!!!!!! I''m speechless~ where did you see/hear about this story?
    
    \textit{Example 2} Just updated my MySpace profile~ am about to send Sweet Danny a comment on his MySpace page~ going to sleep shortly~ zzzzzzz!!!!! 
    
    \textit{Example 3} Had fun; happy day~ going to sleep for Church; Sunday School in the morning~ ZZZZZZZZZZZZZZZZZZZZZ!!!!!!!!!!!!!!!!! 
    
    \textit{Example 4} Am doing wash\~ and as always\~ listening to Danny!!!!!!!!!!!!!!!!!!!!!!!!!! 
    
    \textit{Example 5} @dannygokey Happy to hear you're working on Sophia's Heart~ I'll always support you; Sophia's Heart!!!!!!!!!!!!! \newline
    
    \textit{Test Input} @Cupcake1012 I <3 the Gokey Gang; Danny always; forever~ Danny has the best fans ever!!!!!!!!!!!!!!!!!!!!!!!!!!!!!!!!!  \\
    \midrule
    User 2 \newline (Sent140, \newline Artifacts) &  
    \textit{Example 1} ... A kidney stone. Really?? Ugh \url{http://tinyurl.com/qsw9vq}
    
    \textit{Example 2} Ahhh everything hurts... warming up my bed buddy and going back to sleep. \url{http://tinyurl.com/ol4ugp}
    
    \textit{Example 3} Feelin so sick its dumb. \url{http://tinyurl.com/o6glqr}
    
    \textit{Example 4} I got so much done today!! Got my car fixed, went to the bank. Now im at work to make monies.\url{ http://tinyurl.com/of2ane}
    
    \textit{Example 5} name the movie... ``i wanna be like yooo-oo-ooo....''; \url{http://tinyurl.com/pqbyst}\newline
    
    \textit{Test Input} i wish i was a snail  \url{http://tinyurl.com/pgysd5} \\
    \midrule 
    User 3 \newline (Reddit, \newline ``Justice'' topic) &  
    \textit{Example 1}  dion lewis stiff arm plesse my favorite professor in law school actually argued at scotus
    
    \textit{Example 2} scotus for the petitioner in mccleskey i hope for his
    
    \textit{Example 3} his sake ( and for the sake of justice ) that
    
    \textit{Example 4} that they revisit it at some point \newline
    
    \textit{Test Input} ... am a personal injury attorney <PAD>\\
    \bottomrule
    \end{tabular}
    \normalsize
    \caption{Examples reflecting consistent linguistic patterns, artifacts, and topics across a user's examples.}. 
    \label{tab:userexamples}
    \end{center}
\end{table*}

\paragraph{User vs. Non-User Level Task Demonstrations} In Section \ref{sec:experiments}, we observe that the ``User Privacy'' prompts provide larger performance gains compared to the ``No User Privacy'' prompts. We hypothesize that \textbf{(1) label distributions} user level task demonstrations help because user labels are generally in a restricted subset of classes from the overall task, and \textbf{(2) input distributions} user level task demonstrations help due to linguistic or topical similarity between user demonstrations.

Towards the former, on a per-user basis, as the entropy of the training data labels decreases, the accuracy tends to increase, as in Figure \ref{fig:whyuserlevel}. The entropy is computed for users with at least $4$ training examples and each point represents at least $200$ examples. As the entropy of the user-level train set increases, the improvement of the ``User Privacy'' baseline over the ``No User Privacy'' baseline tends to decrease, suggesting this is an important reason for the difference, though there is still some gap between the baselines for high entropy points. Towards the latter, qualitatively certain users clearly write with specific artifacts or about certain topics, though overall the examples are quite noisy. Three users for whom this occurs are in Table \ref{tab:userexamples}.


\begin{figure*}
    \centering
    \includegraphics[width=0.49\linewidth]{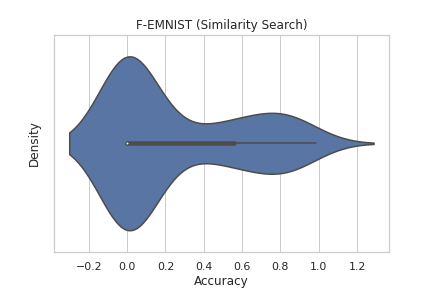}
    \includegraphics[width=0.5\linewidth]{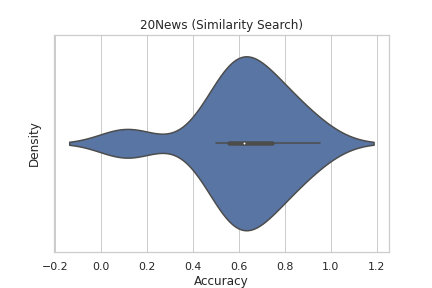}
    \caption[width=\linewidth]{Violin plots by class for the fine-grained classification tasks. The plots are multi-modal suggesting the FMs give unequal zero-shot performance across classes.}
    \label{fig:violinclass}
\end{figure*}

\begin{figure*}
    \centering
    \includegraphics[width=0.49\linewidth]{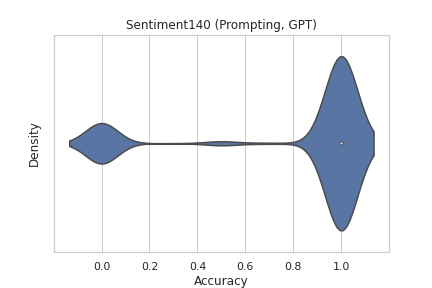}
    \includegraphics[width=0.49\linewidth]{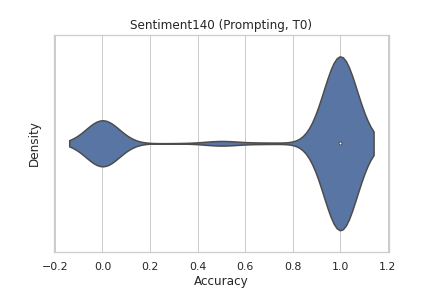}
    \includegraphics[width=0.49\linewidth]{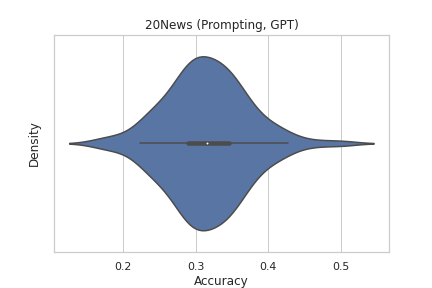}
    \includegraphics[width=0.49\linewidth]{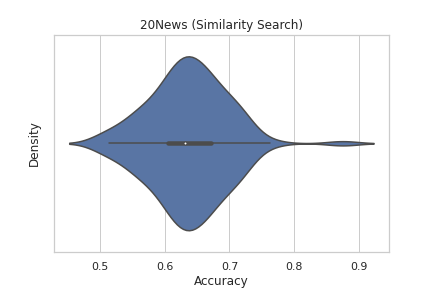}
    \includegraphics[width=0.49\linewidth]{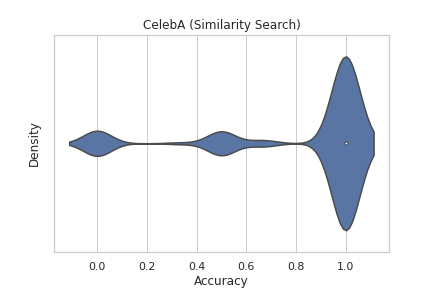}
    \includegraphics[width=0.49\linewidth]{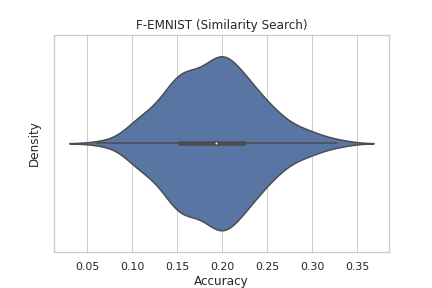}
    \includegraphics[width=0.49\linewidth]{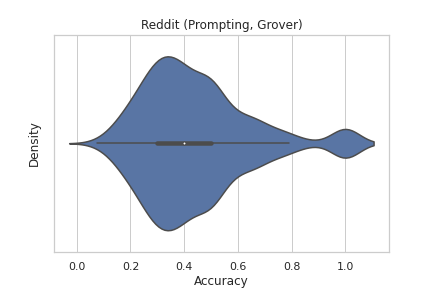}
    \includegraphics[width=0.49\linewidth]{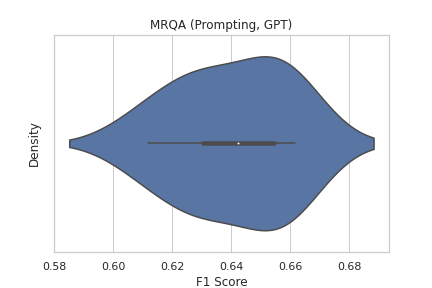}
    \caption[width=\linewidth]{Violin plots by user id. We note that the Sent140, MRQA, CelebA, and Reddit are Non-IID and F-EMNIST and 20News use IID mappings of data to clients. Non-IID plots appear more multi-modal.}
    \label{fig:violin}
\end{figure*}

\paragraph{Violin Plots} Figures \ref{fig:violin} and \ref{fig:violinclass} provide violin plots with respect to zero-shot performance across tasks. These plots demonstrate that on tasks where user data are non-IID, the model performs unevenly in zero-shot use across different groups. Further, on the fine-grained classification tasks, performance is uneven across classes. For example, on F-EMNIST, the model often \textit{either} always selects the uppercase or lowercase version of a letter, resulting in close to 100\% accuracy on the selected case and 0\% on the unselected case.
On 20News, accuracy by class is above 50\% for every class except for ``politics'', where the accuracy is 12.9\% and ``religion'', where the accuracy is 9.6\%. We suggest this is because ``politics guns'' and ``politics in the middle east'' related to politics, and ``Christianity'' and ``athiesm'' related to religion are other classes in the 20News task --- using the inference-only strategies based on natural language descriptions of these classes, it is difficult to express the distinction between these similar classes. We require additional investigation into the zero-shot robustness of FMs, and effective ways of describing classes that avoid class embeddings that point in the same or similar directions in embedding space.

\label{sec:appendix_baselines}

\subsection{Additional Discussion of Baselines}
Given the heterogeneity of FL evaluations \cite{chaoyanghe2020fedml}, we aim to compare \textit{vanilla} baselines for both paradigms in this first investigation. For baselines within the \systemname{} and FL frameworks, we consider advanced neural architectures with standard training and inference methods. For \systemname{}, this involves off-the-shelf, publicly accessible models with manual prompting, without any tuning. For FL, we include numbers that use the most advanced architecture reported for the benchmark in prior work (to the best of our knowledge), using standard FedAvg for training and \textit{none of the following factors} which significantly degrade quality. 
\begin{itemize}
    \item \textbf{Differential privacy} Differential privacy (DP) seeks to provide a guarantee that we cannot reconstruct or memorize the sensitive examples appearing in our training data. The basic approach to achieve this involves adding noise to the data during preprocessing, gradients during training, or elsewhere in the ML pipeline, where the amount of noise depends on the desired privacy parameters. DP faces a tradeoff between stronger privacy protection (achieved by adding more noise) and better convergence performance \cite{wei2019dpfl}. Our FL baselines applied no DP.
    \item \textbf{Adversarial users} Adversaries can execute a data poisoning attack (i.e. poison the labels of the contributed data), send random updates to the central server, and/or execute a model replacement attack (i.e. manipulate the shared model to enforce that it performs a desired subtask, while maintaining performance on the original FL task) \cite{bagdasaryan2019flpoison, xu2020fairfl}. As the fraction of adversarial participants increases, the quality of the FL model tends to degrade \cite{li2021ditto}.
    
    \item \textbf{Adversarial central server} Because the central server is a single point of failure and attacks upon the central server significantly compromise privacy, decentralized federated learning has emerged as an alternative line of work. However, this again degrades quality \cite{sun2021decentralizedfl}.
    \item \textbf{Communication efficiency} Popular methods to reduce the communication cost of FL such as compressing the gradients passed between client and server tend to degrade quality \cite{rothchild2020fetchsgd}.
    \item \textbf{Non-IID Data} \textit{Note that some of our benchmarks are already non-IID}. Under SGD, we require an unbiased estimate of the full gradient. This relies on computing the stochastic gradients over IID samples. User data is likely non-IID, which can degrade the quality and increase the convergence time of the FL model trained with FedAvg \cite{zhao2018noniid}. However, fortunately recent architectures appear more robust to heterogeneous data under FL \cite{qu2022vitfl}.
\end{itemize}
A natural follow up is to compare implementations under \systemname{} to FL under these threats.

\newpage
\vspace{-3mm}

\end{document}